\documentclass[10pt]{article} 

\usepackage[preprint]{rlj} 

\usepackage{amssymb}            
\usepackage{mathtools}          
\usepackage{mathrsfs}           
\usepackage{graphicx}           
\usepackage{subcaption}         
\usepackage[space]{grffile}     
\usepackage{url}                
\usepackage{lipsum}             

\usepackage{amsthm}

\title{Evolution of Societies via Reinforcement Learning}

\setrunningtitle{Evolution of Societies via Reinforcement Learning}

\author{Yann Bouteiller, Karthik Soma, Giovanni Beltrame}

\contribution{
    This paper introduces a methodology for analyzing the social dynamics that stem from MARL revision protocols in large populations.
    }
    {
    While Evolutionary Game Theory traditionally models biological evolution through fitness-based replication in pairwise interactions—an approach later extended to study imitation in social systems—our work explores how societies evolve when agents actively optimize their own fitness via continual learning. This framework is relevant for understanding the dynamics of competitive environments like economic markets.
    }

\contribution{
    We derive fast implementations of exact multi-agent Policy Gradient and exact Opponent-Learning Awareness targeting evolutionary simulations in pairwise matrix games.
    }
    {
    The scope of our paper is limited to non-repeated games with random agent pairing at each evolutionary step. This foundational work establishes a framework that future research can extend to more complex scenarios, including episodic MARL environments (such as the Iterated Prisoner's Dilemma) and structured assortment processes.
    }

\contribution{
   We conduct population simulations of 200,000  agents to analyze how naive learning and Opponent-Learning Awareness shape collective behavior in evolutionary settings. 
    }
    {
    This work represents the first application of advanced MARL revision protocols in the context and scale of Evolutionary Game Theory simulations.
    }


\summary{
Diverse studies from computational neuroscience have found evidence of Reinforcement Learning (RL) driving learning in biological brains, amongst other learning protocols such as imitation.
Furthermore, RL algorithms have now emerged as the core technique explicitly driving innovation in a growing number of industrial applications, including artificial language generation, drug discovery and finance.
It is therefore natural to consider Multi-Agent Reinforcement Learning (MARL) as one of the revision protocols powering social dynamics.
However, the mathematical complexity of this idea has prevented the development of a theory around it so far.
In this paper, we hope to initiate this line of research by leveraging simulation.
We develop computationally efficient implementations of two fundamental MARL revision protocols: "naive" Policy Gradient and Learning with Opponent-Learning Awareness (LOLA). These implementations enable us to conduct large-scale evolutionary simulations across classic interaction models from Evolutionary Game Theory. 
Our experiments yield various insights into how non-stationarity-aware learners affect the evolution of societies.
In particular, we find that LOLA learners promote cooperation in the Stag Hunt model, delay cooperative outcomes in the Hawk-Dove model, and reduce strategy diversity in the Rock-Paper-Scissors model.

}

\begin{document}

\makeCover  

\begin{abstract}
The universe involves many independent co-learning agents as an ever-evolving part of our observed environment.
Yet, in practice, Multi-Agent Reinforcement Learning (MARL) applications are typically constrained to small, homogeneous populations and remain computationally intensive.
We propose a methodology that enables simulating populations of Reinforcement Learning agents at evolutionary scale.
More specifically, we derive a fast, parallelizable implementation of Policy Gradient (PG) and Opponent-Learning Awareness (LOLA), tailored for evolutionary simulations where agents undergo random pairwise interactions in stateless normal-form games.
We demonstrate our approach by simulating the evolution of very large populations made of heterogeneous co-learning agents, under both naive and advanced learning strategies.
In our experiments, 200,000 PG or LOLA agents evolve in the classic games of Hawk-Dove, Stag-Hunt, and Rock-Paper-Scissors.
Each game provides distinct insights into how populations evolve under both naive and advanced MARL rules, including compelling ways in which Opponent-Learning Awareness affects social evolution.
\end{abstract}

\begin{figure}[htbp]
    \centering
    \begin{subfigure}[b]{0.32\textwidth}
        \centering
        \includegraphics[width=\textwidth, trim=0 0 0 0]{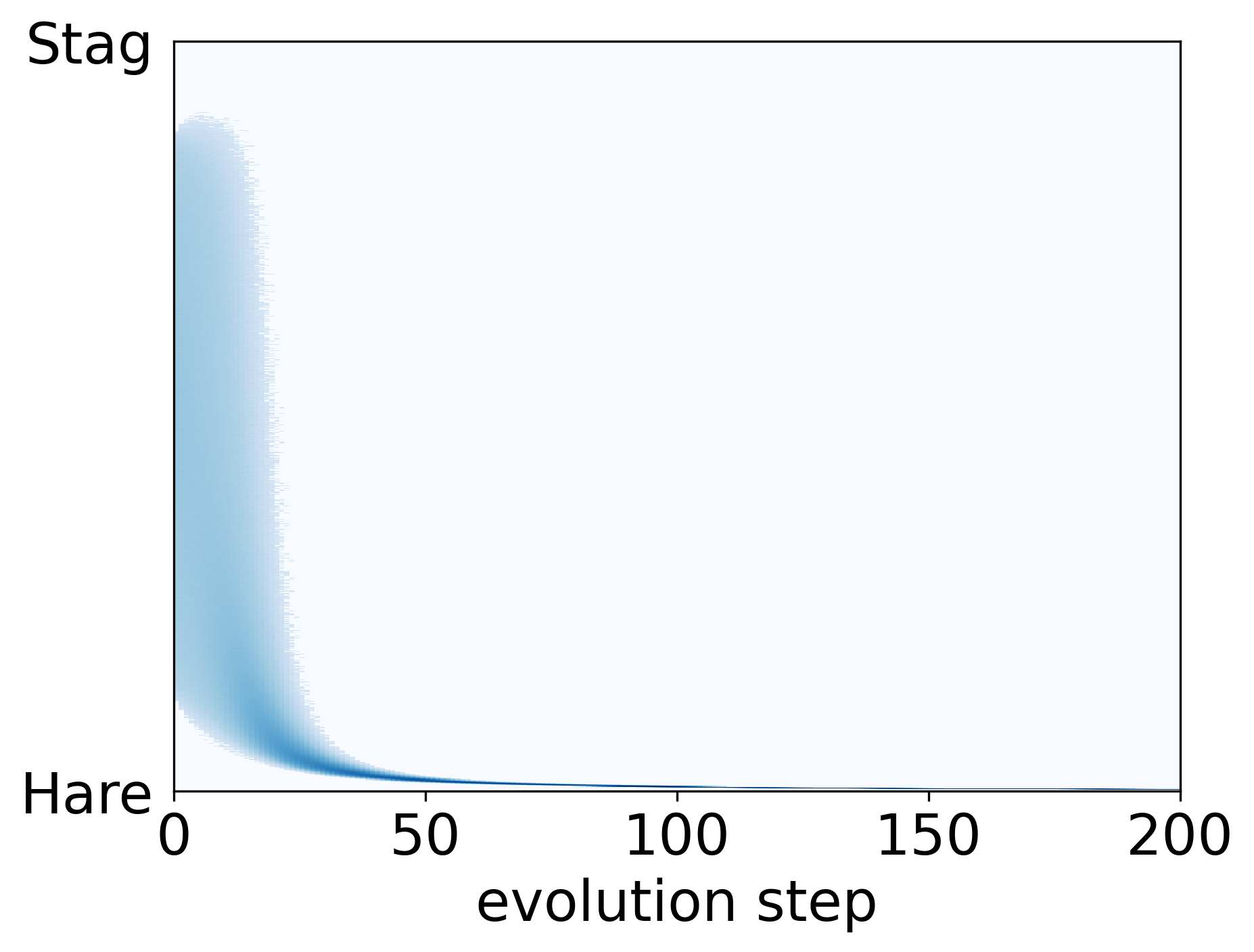}
        \caption{PG in SH ($s=1.8$)}
        \label{fig:sh_evo_pg}
    \end{subfigure}
    \hfill
    \begin{subfigure}[b]{0.32\textwidth}
        \centering
        \includegraphics[width=\textwidth, trim=0 0 0 0]{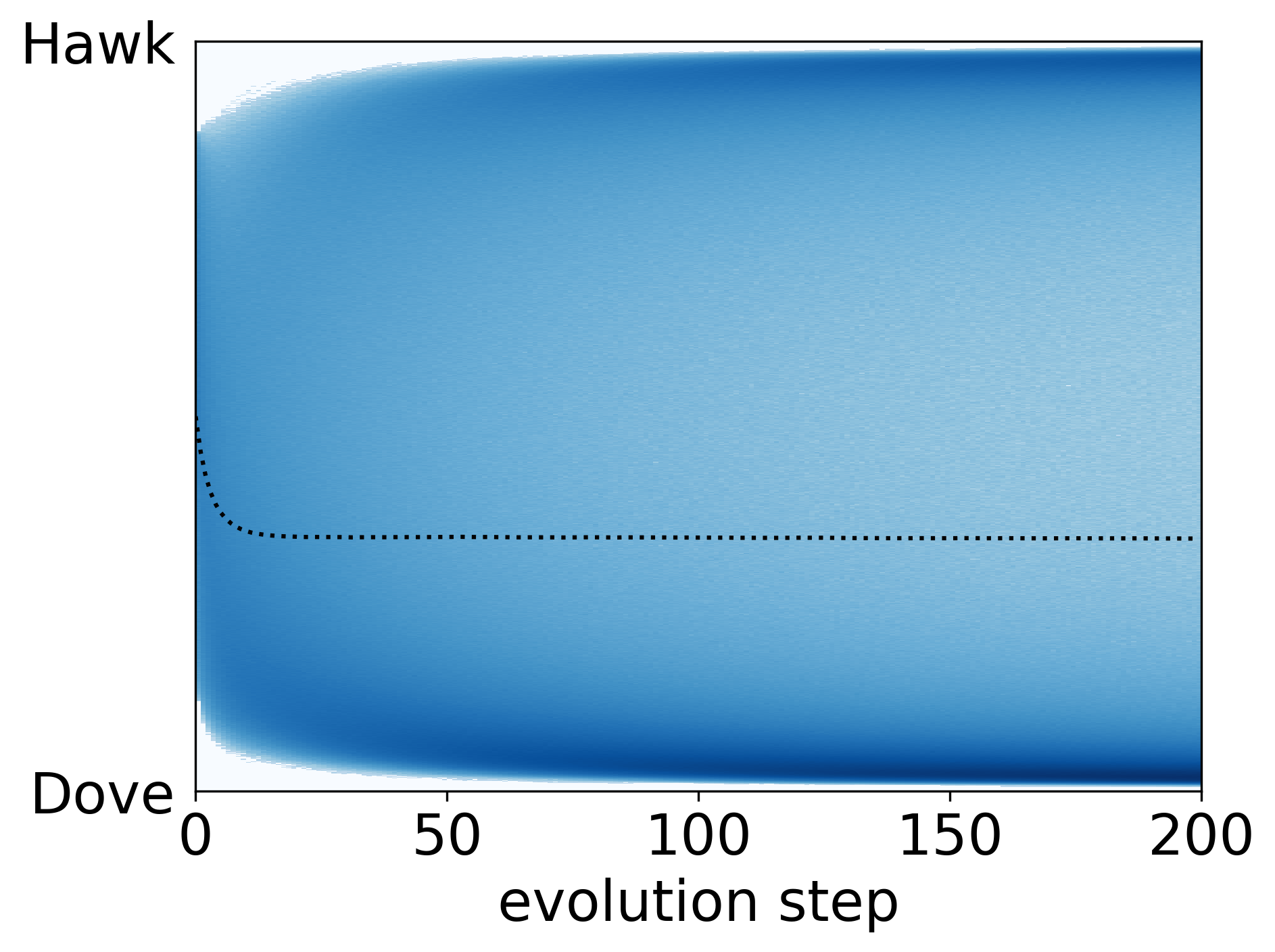}
        \caption{PG in HD ($f=-2$)}
        \label{fig:hd_evo_pg1}
    \end{subfigure}
    \hfill
    \begin{subfigure}[b]{0.32\textwidth}
        \centering
        \includegraphics[width=\textwidth, trim=100 50 60 50, clip]{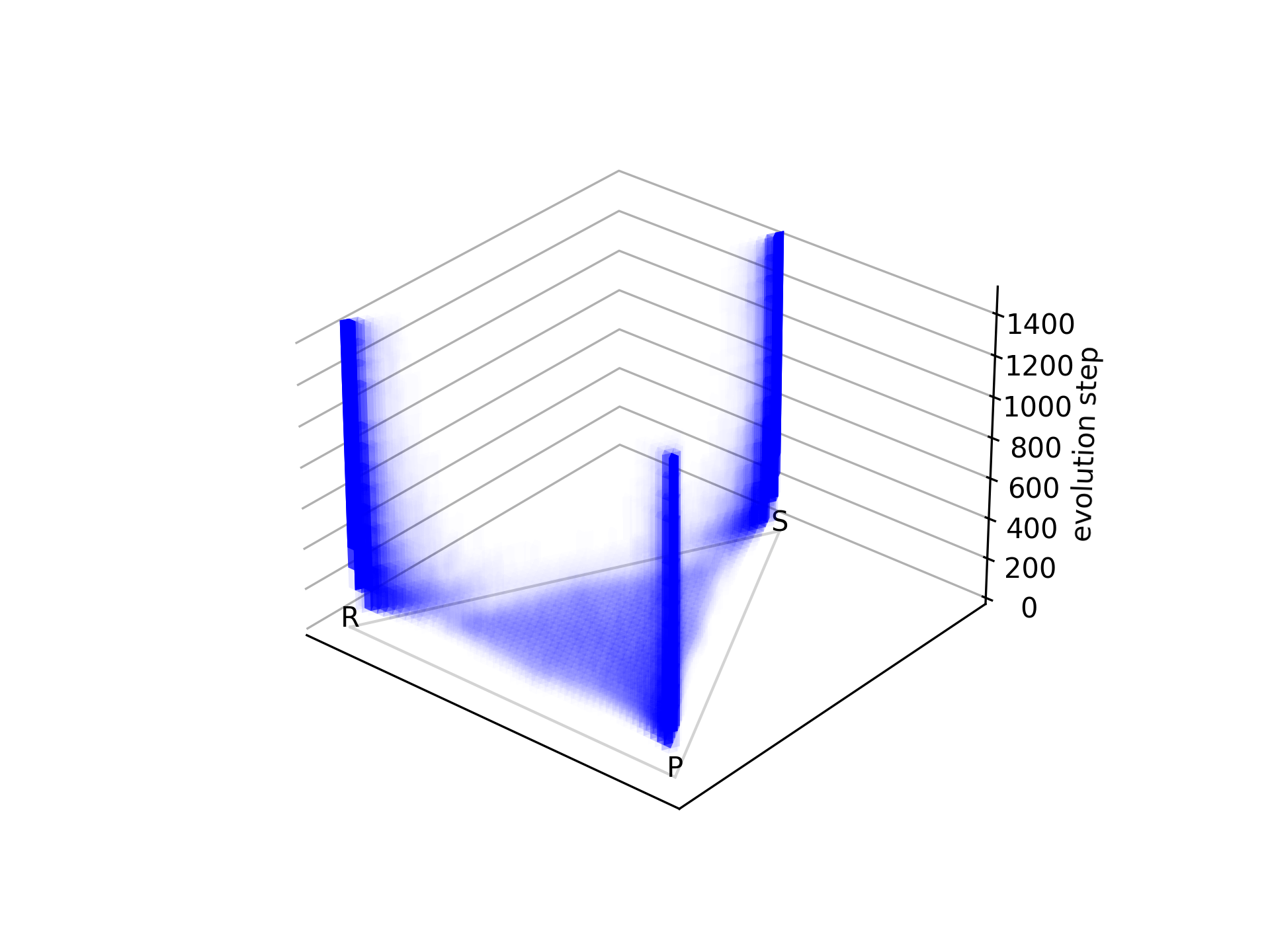}
        \caption{PG in RPS}
        \label{fig:rps_evo_pg}
    \end{subfigure}
    \hfill
    \begin{subfigure}[b]{0.32\textwidth}
        \centering
        \includegraphics[width=\textwidth, trim=0 0 0 0]{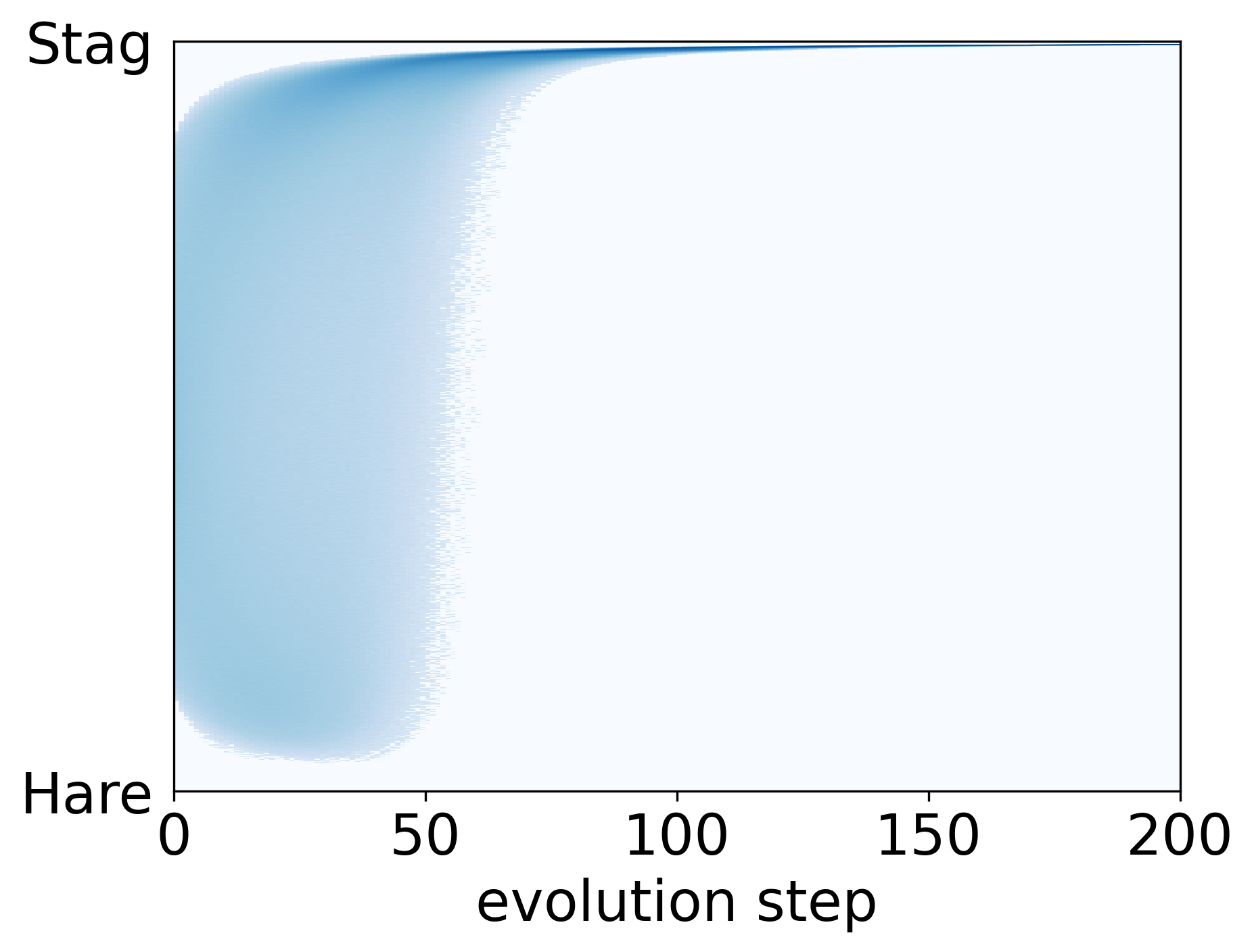}
        \caption{LOLA in SH ($s=1.8$)}
        \label{fig:sh_evo_lola}
    \end{subfigure}
    \hfill
    \begin{subfigure}[b]{0.32\textwidth}
        \centering
        \includegraphics[width=\textwidth, trim=0 0 0 0]{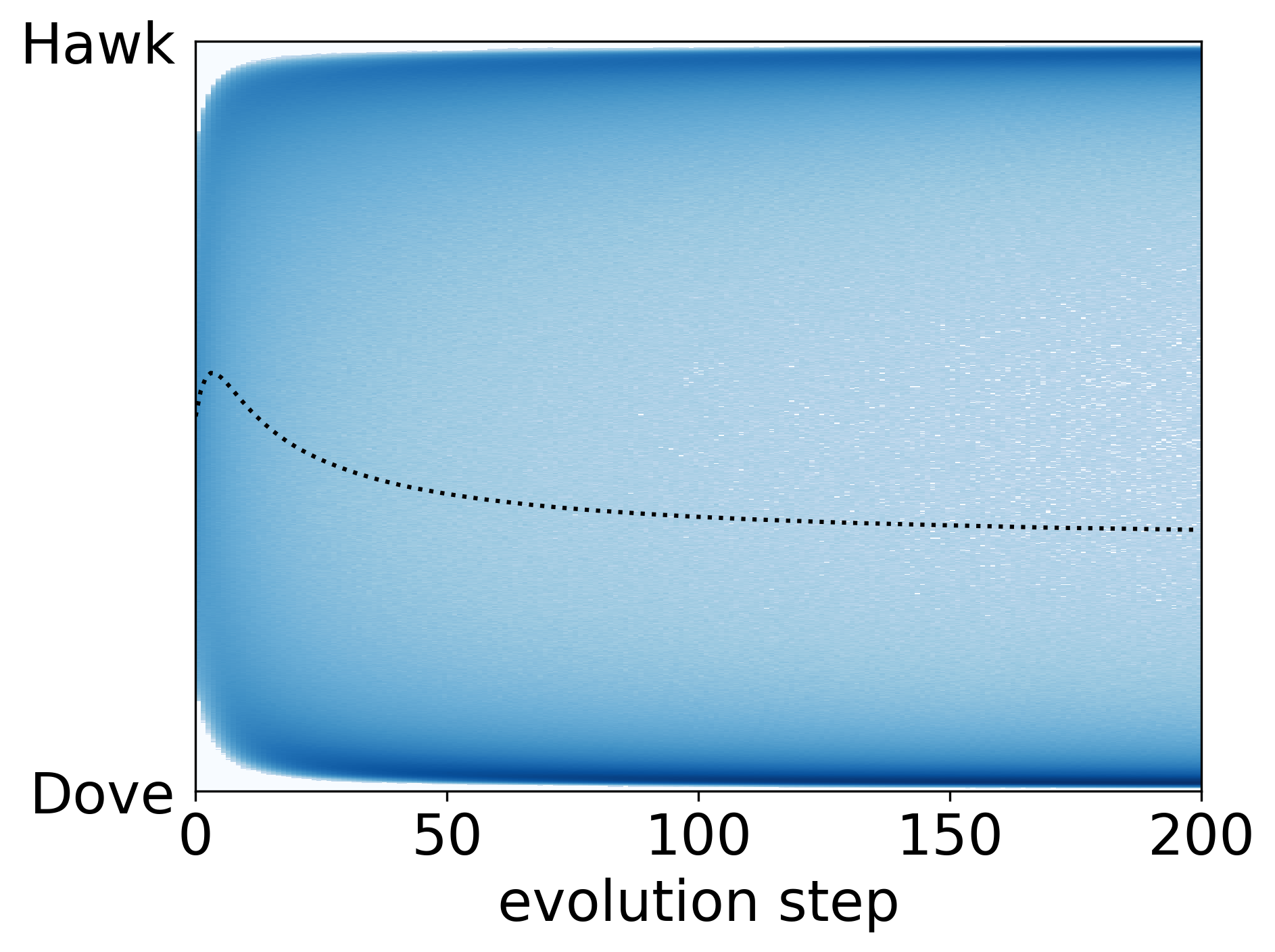}
        \caption{LOLA in HD ($f=-2$)}
        \label{fig:hd_evo_lola1}
    \end{subfigure}
    \hfill
    \begin{subfigure}[b]{0.32\textwidth}
        \centering
        \includegraphics[width=\textwidth, trim=100 50 60 50, clip]{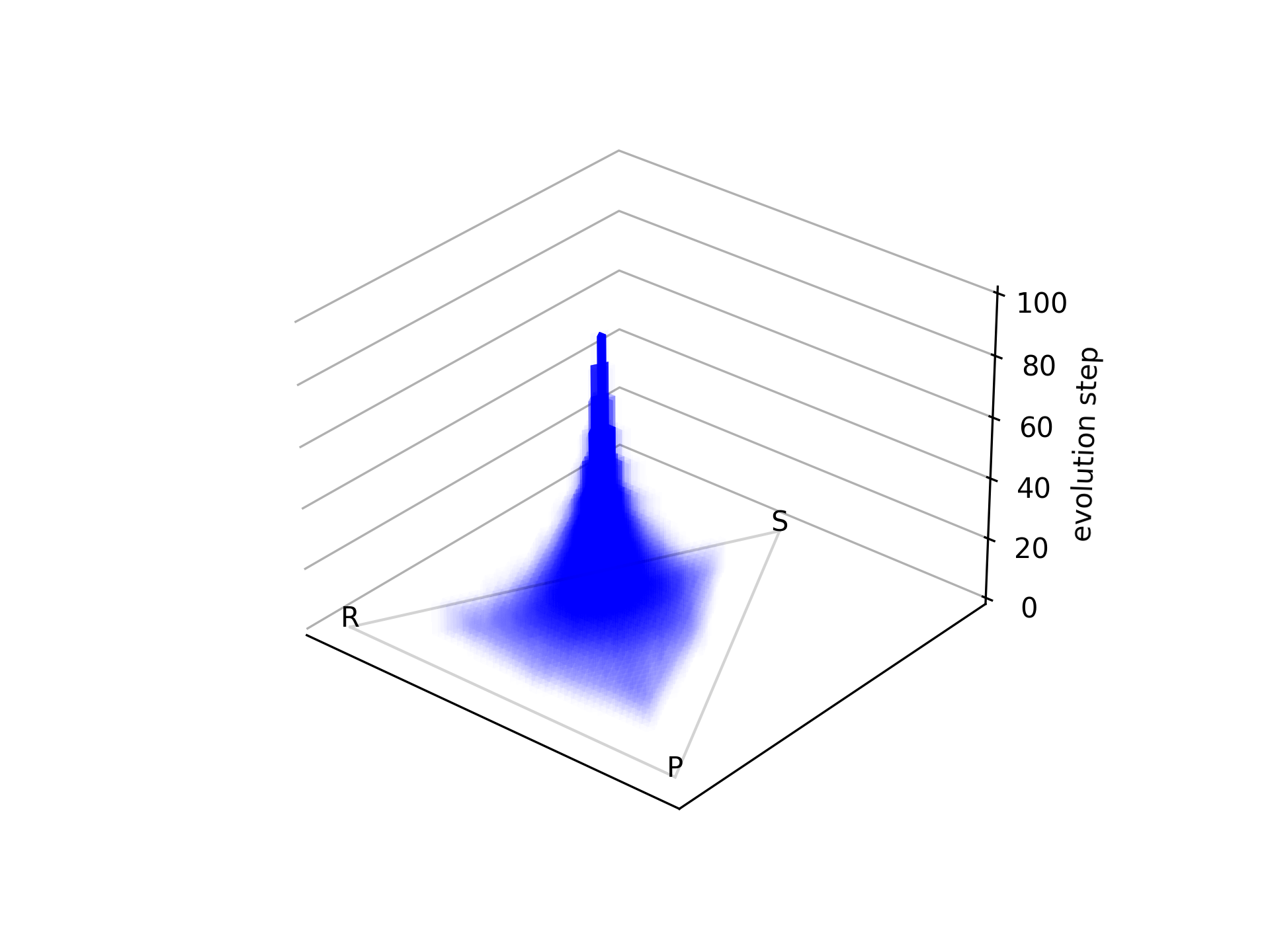}
        \caption{LOLA in RPS}
        \label{fig:rps_evo_lola}
    \end{subfigure}
    \caption{Populations of 200,000 RL agents evolving in the classic games of Stag Hunt, Hawk-Dove and Rock-Paper-Scissors (columns), via Policy Gradient and LOLA (rows). Each agent is a stochastic policy, represented as linear coordinates between pure strategies. Dark shades of blue indicate high concentrations of agents, and evolution steps correspond to one learning step performed per agent. In Hawk-Dove, black dots indicate the average policy over the entire population.}
    \label{fig:all_evo}
\end{figure}

\begin{figure}[htbp]
    \centering
    \begin{subfigure}[b]{0.32\textwidth}
        \centering
        \includegraphics[width=\textwidth, trim=0 0 0 0]{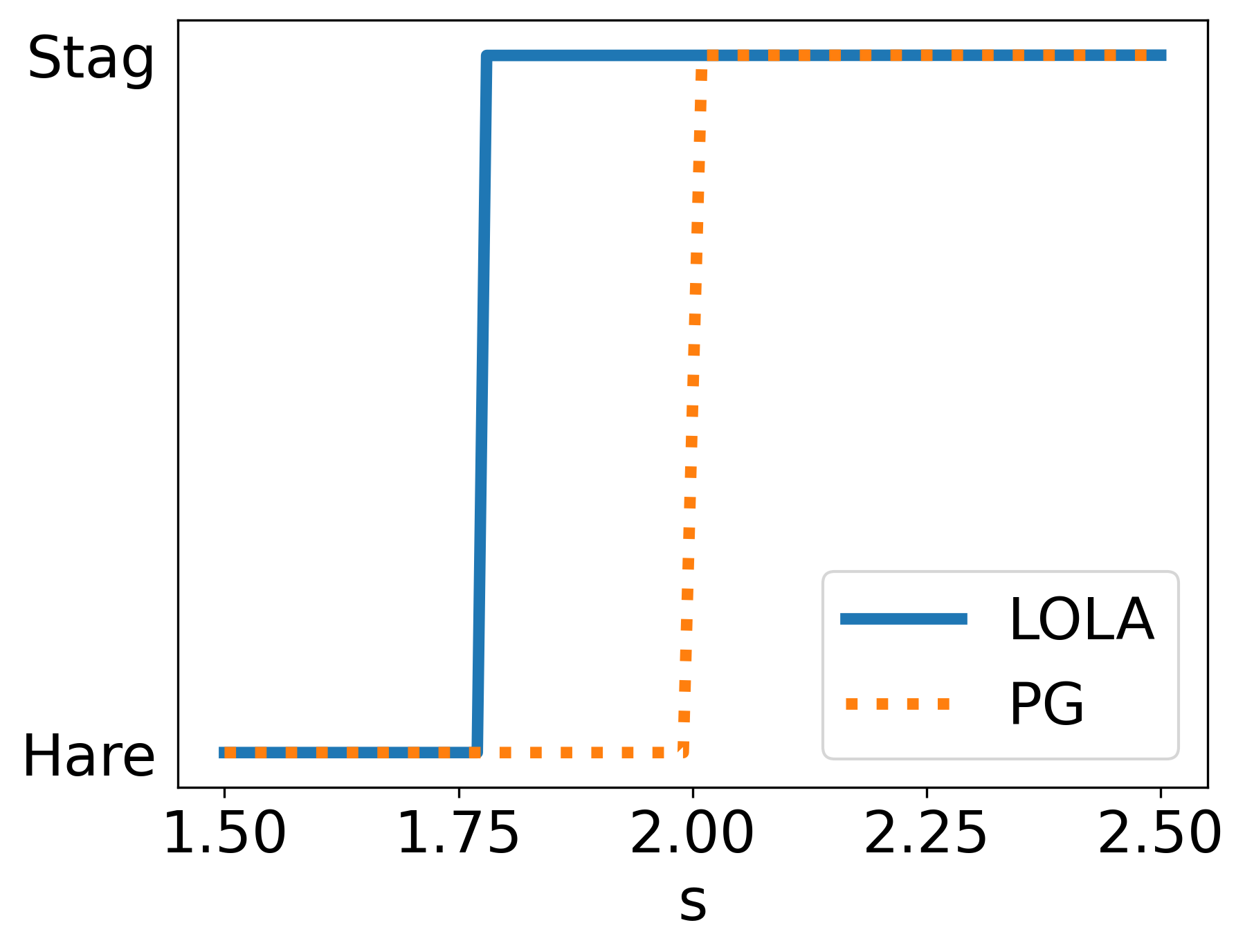}
        \caption{Stag Hunt}
        \label{fig:sweep_sh}
    \end{subfigure}
    \quad
    \begin{subfigure}[b]{0.32\textwidth}
        \centering
        \includegraphics[width=\textwidth, trim=0 0 0 0]{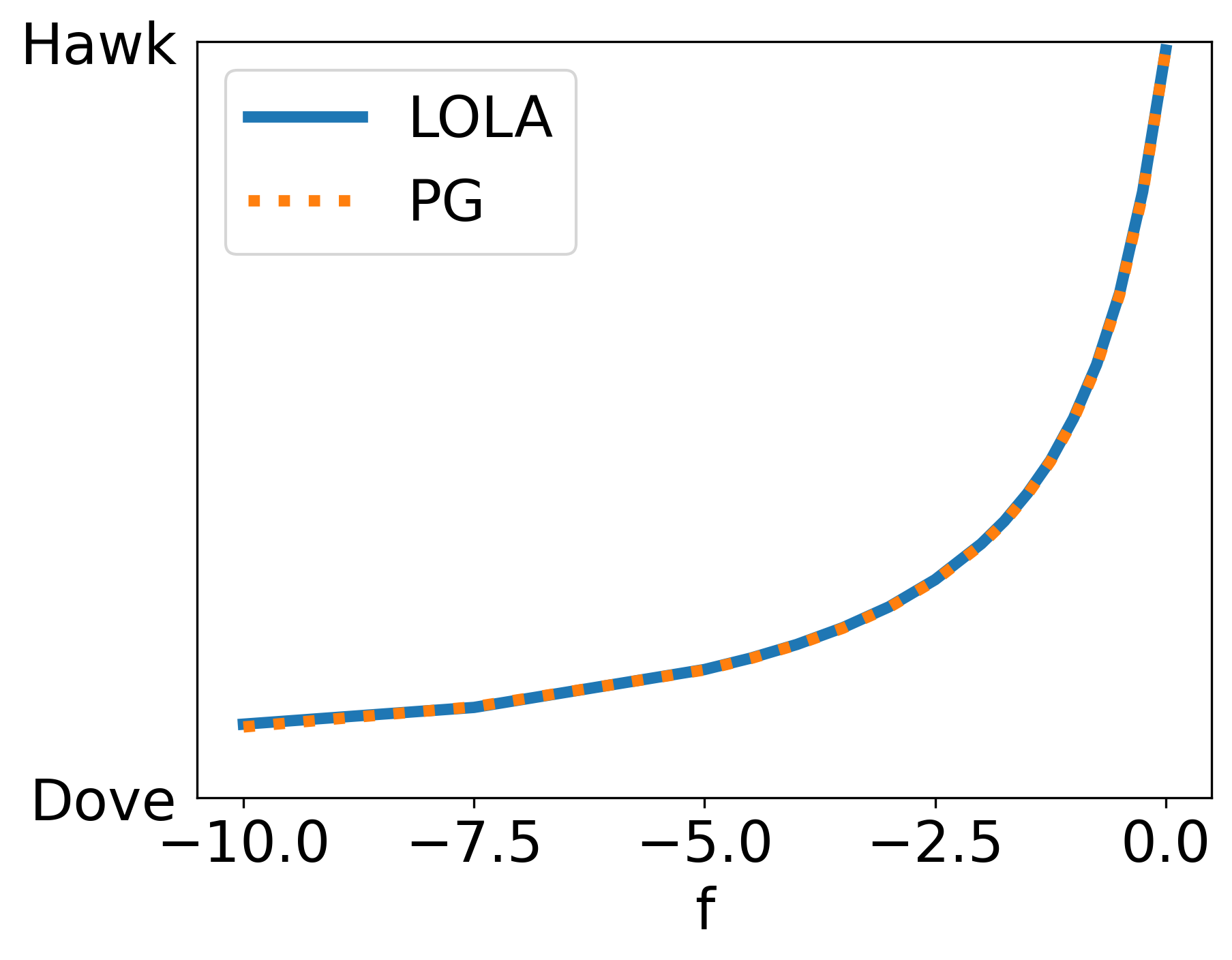}
        \caption{Hawk Dove}
        \label{fig:sweep_hd}
    \end{subfigure}
    \caption{Final average policy over the population, depending on cost values.}
    \label{fig:sweeps}
\end{figure}

\section{Introduction}

Our universe is one of perpetual change, where countless agents co-exist, co-learn and co-evolve.
From an individual agent's perspective, other learning agents are a fundamentally non-stationary part of the environment, especially  when incentives are in conflict \citep{papoudakis2019nstationary}.
In the realm of Reinforcement Learning (RL), the study of this type of possibly adversarial non-stationarity is a field known as Multi-Agent Reinforcement Learning (MARL).
Some notable achievements do tackle MARL scenarios, but these are largely constrained by the high complexity of multi-agent training.
In fact, these often resort to ``self-play''~\citep{silver2016mastering, berner2019dota}, i.e., training a single neural network against one or few copies of itself, or to ``centralized training''~\citep{lowe2020maddpg, yu2022mappo}, i.e., circumventing non-stationarity by using privileged global information.

We are interested in modeling plausible real-world social evolution processes that stem from RL revision protocols.
We make the assumption that individuals continuously adapt their own strategies by following local RL rules, and we simulate how large societies evolve as a result.
This setting is relevant to the study of social dynamics where individuals actively optimize their own fitness, such as economy and finance.
More precisely, the scope of this paper is to simulate large populations of persistent RL agents, following an interaction model close to that of Evolutionary Game Theory (EGT).
At each evolution step, agents are paired according to some assortment process (in our case, uniformly at random), and interact according to their respective policies.
After each such interaction, agents adapt their policies according to their respective payoff and learning rule\footnote{Throughout this paper, we use vocabulary from RL and EGT interchangeably. In particular, ``payoff'' = ``return'', ``fitness''~= ``value'', ``strategy'' = ``policy'', ``pure strategy'' = ``action'', ``revision protocol'' = ``learning rule''.}.
Note that this type of social evolution is intrinsically driven by learning and individual preferences, in stark contrast to genetic evolution, which the literature usually considers as an extrinsic process happening among non-learning agents through their replication and death~\citep{weibull1997evolutionary}.
In our work, the concept of \emph{fitness} does not necessarily refer to a measure of reproductive ability. Instead, it refers to a measure of expected completion of an arbitrary individual objective, i.e., an RL value.

We bring advanced intrinsic MARL protocols to the scale of evolutionary simulations, studying in particular the evolutionary effects of Opponent-Learning Awareness, an advanced MARL protocol able to take advantage of non-stationary dynamics~\citep{foerster2017learning}.


\section{Preliminaries}

This section reviews the concepts from Evolutionary Game Theory (EGT) and Multi-Agent Reinforcement Learning (MARL) used in this paper.
A high-level overview of the literature contextualizing our work is provided in the Supplementary Material (Appendix~\ref{ap:related}).

\subsection{Evolutionary Game Theory (EGT)}\label{sec:egt}

EGT models evolution as a series of random pairwise interactions, where interactions are typically simple bi-matrix games (i.e., 2-player multi-armed bandits).
Agents are sampled from a large population to be randomly paired and evaluated against their drawn opponent. 
The outcome of this interaction is a payoff for each opponent, whose expectation is called the agent's \emph{fitness} against the current population.
An agent's fitness depends on both its policy and the current configuration of the population.
The agent's policy is called its \emph{type}, which is one of $n$ possible types.
In gene-inspired revision protocols, agents with a greater fitness replicate and thus tend to ``invade'' the population, whereas agents with a lower fitness tend to go ``extinct''.
In particular, EGT is interested in \emph{evolutionarily stable equilibria}, which are configurations of the population where the different types are present in stable proportions under replication dynamics.
In evolutionarily stable equilibria, the population configuration is robust to rare mutations, where few agents randomly switch from one type to another.
In this paper, we will be studying more complex, learning-based evolutionary equilibria in three classic symmetric games: \emph{Stag Hunt}, \emph{Hawk-Dove}, and \emph{Rock-Paper-Scissors}.
These games can be described by bi-matrices:

\begin{table}[ht]%
   \centering
  \subfloat[tab:sh][Stag Hunt]{
   $
    \begin{array}{r|cc}
      & \text{Stag} & \text{Hare} \\
      \hline
      \textbf{Stag} & s,s & 0,1 \\
      \textbf{Hare} & 1,0 & 1,1 \\
    \end{array}
    $
  }%
  \qquad
  \subfloat[][Hawk-Dove]{
  $
    \begin{array}{r|cc}
  & \text{Hawk} & \text{Dove} \\
  \hline
  \textbf{Hawk} & f,f & 2,0 \\
  \textbf{Dove} & 0,2 & 1,1 \\
\end{array}
    $
  }
  \qquad
  \subfloat[][Rock-Paper-Scissors]{
  $
    \begin{array}{r|ccc}
  & \text{Rock} & \text{Paper} & \text{Scissors} \\
  \hline
  \textbf{Rock} & 0,0 & -1,1 & 1,-1 \\
  \textbf{Paper} & 1,-1 & 0,0 & -1,1 \\
  \textbf{Scissors} & -1,1 & 1,-1 & 0,0 \\
\end{array}
    $
  }
  \label{tbl:table}%
\end{table}
where rows represent the action chosen by the ego agent (bold) with corresponding payoffs in first position, and columns represent the action chosen by the other agent with corresponding payoffs in second position.

\textbf{Stag hunt} (SH) is a 2-action game illustrating the evolution of cooperation.
In this game, agents need to hunt for food and choose to either go for a Stag, or go for a Hare.
Hunting a Hare is easy: any agent choosing this option successfully receives a payoff of $1$.
Hunting a Stag is harder: both agents need to cooperate, otherwise the agent choosing to go for a Stag fails to catch anything and receives a payoff of $0$.
However, if both agents cooperate, they succeed and each receives a payoff of $s > 1$, which is better than going for Hares.
The game of Stag hunt has two distinct pure strategy Nash equilibria\footnote{2-player equilibria where each agent always selects the same action.}: (1) both agents always playing Stag, and (2) both agents always playing Hare.

\textbf{Hawk-Dove} (HD) is a 2-action game illustrating the evolution of conflict over shareable resources.
Agents either choose to act as a ``Hawk'' or as a ``Dove''.
When a Dove encounters another Dove, they share the available food (each receives a payoff of $1$).
When a Dove encounters a Hawk, it yields and gets no food (payoff of $0$) while the Hawk gets all of it (payoff of $2$).
But when a Hawk encounters another Hawk, they fight and both get injured (payoff of $f<0$).
The game of Hawk-Dove has two pure strategy Nash equilibria: (1) Hawk-Dove and (2) Dove-Hawk.
But note that in these equilibria, Doves have a smaller payoff than Hawks.
In the context of evolutionary genetics, this means that when Doves encounter almost only Hawks, they move toward extinction as Hawks invade.
However, when Hawks encounter almost always Hawks, their expected payoff is even less than Doves encountering Hawks, and thus Hawks move toward extinction as Doves invade.
In other words, there are population configurations in which it is not more relevant to be a Hawk than a Dove in terms of fitness, and replication dynamics naturally drive the population there.

\textbf{Rock-Paper-Scissors} (RPS) is a 3-action zero-sum\footnote{The sum of the two agents' payoffs is always $0$.} game.
It illustrates more complex situations where there are cycles in the preferences over actions.
Scissors beats Paper, Paper beats Rock, and Rock beats Scissors.
RPS has a single mixed-Nash equilibrium, where both players choose their actions uniformly at random.
Similarly to the HD game, all populations whose average behavior is this equilibrium are neutrally stable under replication dynamics and drifting due to random mutations.

A population whose individuals are distributed amongst $n$ types can be represented as a \emph{population vector} $P \in \mathbb{R}^n$ whose components $0 \leq p_i \leq 1$ sum to $1$ and represent the proportion of type $i$.
Under the ``imitation of the fittest'' revision protocol (as well as several other bio-inspired revision protocols), large populations are known to follow a famous population dynamic over time ($t$), called the \emph{Replicator Dynamic}:
\begin{equation}\label{eq:replicator}
    \frac{d p_i}{dt} = p_i (v_i(P) - \bar{v}(P))
\end{equation}
where $v_i(P)$ is the fitness of type $i$ in the population, and $\bar{v}(P)$ is the average fitness of all agents in the population.
Denoting the vector of $v_i$'s as $Q$, the vector of all ones as $\boldsymbol{1}$ and the Hadamard product\footnote{The matrix product takes precedence over the Hadamard product in all our notations.} as $\odot$, Equation~\ref{eq:replicator} can be written in matrix form:
\begin{equation} \label{eq:replicator_mat}
    \frac{d}{dt} P = P \odot (Q - \boldsymbol{1} \bar{v}),
\end{equation}


\subsection{Multi-Agent Reinforcement Learning} \label{sec:marl}

While EGT traditionally focuses on populations where non-learning agents reproduce based on their fitness, our work examines how populations evolve when agents actively learn and adapt their strategies.
Similar to EGT, we model social dynamics as a series of random pairwise interactions.
Therefore, we are principally interested in 2-agent learning rules.
In this paper, we will be specifically looking at two important such learning rules: \emph{policy gradient} (PG), also referred to as ``naive learning'' in the MARL literature, and \emph{learning with opponent-learning awareness} (LOLA) in its full form (i.e., using all terms from the first-order Taylor expansion).

\textbf{Policy gradient} (PG) is a fundamental learning rule from single-agent RL.
It consists of following the first-order gradient of the value function with respect to the ego agent's policy parameters.
Let us consider a pair of agents.
We denote the ego agent as agent $1$, and the other agent as agent $2$.
Let us further denote their respective policies as $\pi_1$ and $\pi_2$, parameterized by vectors $\theta_1$ and $\theta_2$, of current values $v_1$ and $v_2$.
The naive policy gradient is:
\begin{equation}
    \nabla_{\theta_1} v_1(\theta_1, \theta_2)
\end{equation}
The reason why following this gradient is considered naive in MARL is that this does not take into account the non-stationarity introduced by the learning process of agent $2$.

\textbf{Learning with opponent learning awareness} (LOLA) is an improved version of PG that takes into account the learning process of the other agent.
More precisely, LOLA models the learning process of agent $2$ as if agent $2$ were a naive learner and differentiates through its learning step:
\begin{equation}
    \nabla_{\theta_1} v_1(\theta_1, \theta_2 + \Delta \theta_2)
\end{equation}
where
$\Delta \theta_2 = \eta \nabla_{\theta_2} v_2(\theta_1, \theta_2)$
is the naive learning step of agent $2$, $\eta$ being its learning rate.
LOLA approximates this gradient by the following first-order Taylor expansion:
\begin{align} 
\nonumber \nabla_{\theta_1} v_1(\theta_1, \theta_2 + \Delta \theta_2) \approx \nabla_{\theta_1} (v_1 + (\Delta \theta_2)^\top \nabla_{\theta_2} v_1) \\
= \underbrace{\nabla_{\theta_1}v_1}_{\text{PG}} + \underbrace{\eta (\nabla_{\theta_1} \nabla_{\theta_2} v_1)^\top \nabla_{\theta_2} v_2 + \eta (\nabla_{\theta_1} \nabla_{\theta_2} v_2) ^\top \nabla_{\theta_2} v_1}_{\text{opponent-learning awareness}} \label{eq:lola}
\end{align}
Differentiating through the learning step of the opponent has an important advantage in our discussion: it is a naturally plausible way of predicting non-stationarity (assuming we maintain an internal model of others) in order to adapt beforehand and actively steer it toward our own incentives.

\subsection{Population-policy equivalence} \label{sec:equivalence}
In a population of agents playing only pure strategies, 
uniformly sampling agents is equivalent to sampling actions from the abstract stochastic policy defined by the probability vector $P$.
Thus, Equation~\ref{eq:replicator_mat} can be viewed as a learning process, albeit at the population level, where the evolving population of non-learning agents $P$ is itself a self-play learning agent~\citep{bloembergen2015evolutionary}.


\section{Methods}\label{sec:methods}

To model how learning affects societies, we adopt a philosophy similar to EGT.
Namely, we consider large populations of independent learning agents, which are paired randomly at each evolution iteration and interact in normal-form matrix games.
Each agent has its own learning rule (i.e., either of the two presented in Section~\ref{sec:marl}) that it applies to its own policy after each pairwise interaction.
Whereas MARL usually thinks about these rules in the context of persistent interactions between fixed pairs of agents, in the evolutionary setting, they instead get applied after single interactions between random pairs.
In other words, learning agents consider that the random opponent they interact with at each evolution step is a representative sample of the population and, for LOLA, of the current direction of its non-stationary dynamics.

\subsection{Policy architecture}\label{sec:architecture}
From an RL perspective, the normal-form matrix games presented in Section~\ref{sec:egt} are 2-agent multi-armed bandits.
As this is a common assumption in multi-armed bandits \citep{sutton2018reinforcement}, we consider the policy architecture parameterized by the preference vector $\theta \in \mathbb{R}^n$, where $n$ is the number of actions, projected to the probability simplex by a simple softmax function $\sigma$, which yields the probability vector $P \in \mathbb{R}^n$ of the policy selecting each of the $n$ available actions:
\begin{equation} \label{eq:model}
    P = \sigma (\theta)
\end{equation}
This policy architecture has a useful property for our derivations: its gradient has a symmetric analytical form, which is
\begin{equation}
\nabla_{\theta} P = (\nabla_{\theta} P)^\top = \mbox{diag}(P) - P P^\top
\end{equation}

\subsection{Analytical Policy Gradient}\label{sec:analytical_pg}
Let us consider a symmetric normal-form game with $n$ actions, played by a pair of agents denoted as agents $1$ and $2$.
Since the game is symmetric, we can represent its bi-matrix as a single matrix $A \in \mathbb{R}^{n \times n}$, valid from the perspective of both agents\footnote{$A$ is formed by the first entries of the corresponding bi-matrix in Section~\ref{sec:egt}.}.
Let us further assume that the policies of both agents are parameterized by $\theta_{1,2} \in \mathbb{R}^{n}$, with the simple policy architecture described in Equation~\ref{eq:model}: 
\begin{equation}
    P_{1,2} = \sigma (\theta_{1,2})
\end{equation}
The value functions of both agents are:
\begin{equation} \label{eq:value}
    v_1 = P_1 ^\top A P_2 \text{   } ; \text{   }  v_2 = P_2^\top A P_1
\end{equation}
Thus, the ``naive'' Policy Gradient of agent $1$'s value with respect to agent $1$'s parameters is:
\begin{equation*}
    \nabla_{\theta_1} v_1 = P_1 \odot (Q_1 - \boldsymbol{1} v_1)
\end{equation*}
where $Q_1 \in \mathbb{R}^n$ is the vector of action-values of the $n$ available actions (derivation in Appendix~\ref{ap:pg}).

As a side note, this draws an interesting parallel with Equation \ref{eq:replicator_mat}: the PG update on the parameter vector $\theta_1$ is the same as the Replicator update on the probability vector $P_1 = \sigma ( \theta_1 )$. In other words, each individual PG agent can itself be seen as an evolving population of abstract non-learning agents playing pure strategies, similarly to the equivalence noted in \cite{bloembergen2015evolutionary}.

This formulation yields the following analytical PG formulation for symmetric normal-form games:
\begin{align}
    \nabla_{\theta_1} v_1 &= P_1 \odot (I - \boldsymbol{1} P_1^\top) A P_2 \label{eq:pg11} \\
    \nabla_{\theta_2} v_2&= P_2 \odot (I - \boldsymbol{1} P_2^\top) A P_1 \label{eq:pg22}
\end{align}
which only involves simple matrix operations, and therefore is a fast, easily parallelizable implementation.

\subsection{Analytical LOLA}\label{sec:analytical_lola}

Similarly, we derive the following analytical formulation of the LOLA gradient in the symmetric normal-form game defined by matrix $A$ (the derivation is provided in Appendices~\ref{ap:lola} and \ref{ap:fast_lola}):
\begin{align}
\nabla_{\theta_1} v_1(\theta_1, \theta_2 + \Delta \theta_2)
& \approx P_1 \odot X_1 A P_2 \nonumber\notag \\
& +\eta (T^\top \odot X_1 A X_2^\top) (P_2 \odot X_2 A P_1) \nonumber\notag\\
& +\eta (T^\top \odot X_1 A^\top X_2^\top) (P_2 \odot X_2 A^\top P_1) \label{eq:fast_lola}
\end{align}
where $X_1 := I - \boldsymbol{1} P_1^\top$, $X_2 := I - \boldsymbol{1} P_2^\top$, and $T:=P_2 P_1^\top$.
As for Policy Gradient, this formulation only involves simple matrix operations and is straightforward to parallelize.



\subsection{Batched pairwise bandits}\label{sec:batch}
\begin{figure}[h]
    \centering
    \includegraphics[width=0.45\textwidth]{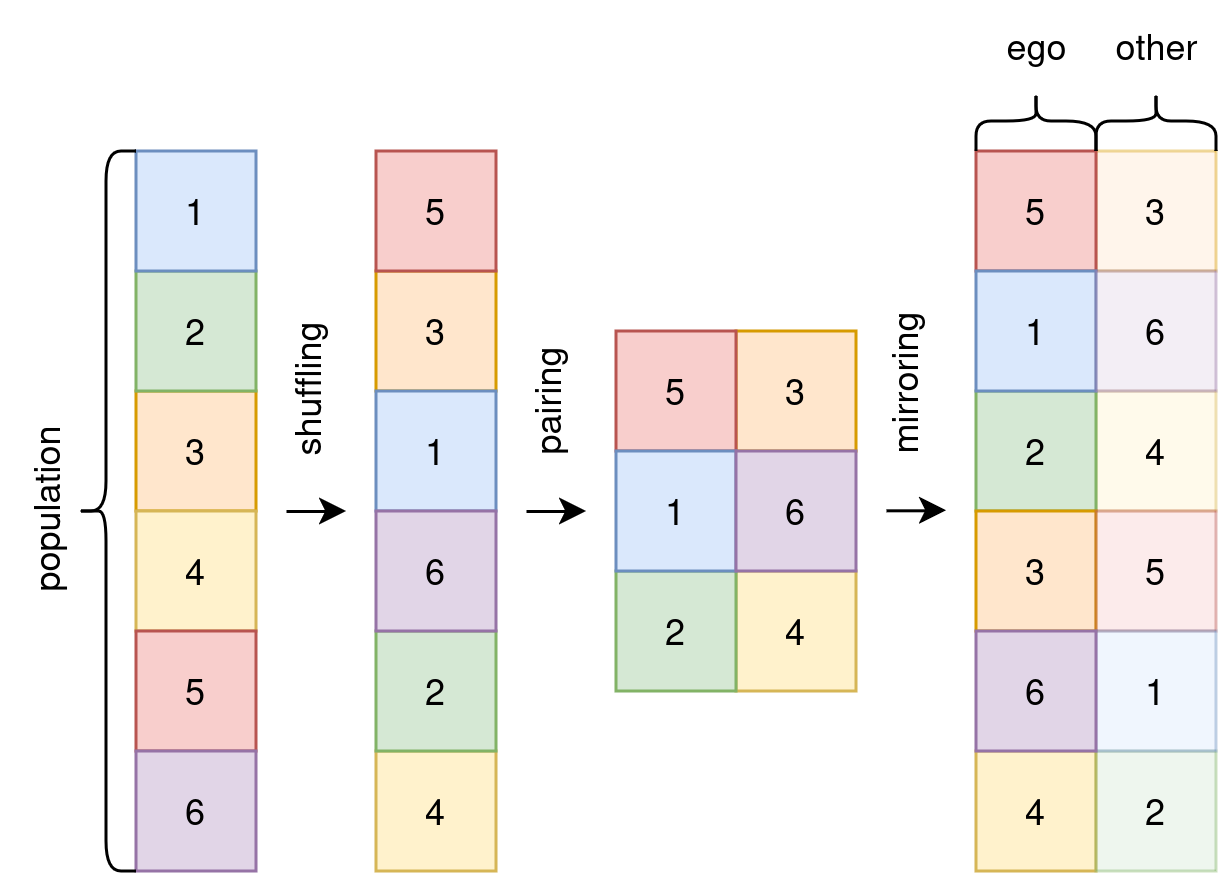} 
    \caption{Pairing and batching}
    \label{fig:batching}  
\end{figure}
Equations~\ref{eq:pg11} and \ref{eq:fast_lola} are fast, backprop-free implementations of exact PG and exact LOLA for normal-form games.
Still, it would be prohibitively slow to apply these updates iteratively on single agent pairs in evolutionary-scale simulations.
At each \emph{evolution step}, all agents go through their revision protocol once.
To make this process scalable, we batch learning updates across the entire population.
More precisely, at each evolution step, we shuffle the entire population and randomly pair all agents two-by-two.
Since all interactions are pairwise in evolutionary simulations, this enables batching updates across agent pairs.
To optimize the process even further, we mirror all pairs to perform the entire population update in one single batched operation.
This batching procedure is illustrated in Figure~\ref{fig:batching}.
We found batching populations in this manner to be extremely efficient.
Combining this procedure with the analytical PG and LOLA implementations described in previous Sections, we are able to simulate populations of 200,000 learning agents for thousands of evolution steps in a matter of seconds on a consumer-grade GPU\footnote{All experiments in this paper are conducted with an i7-12700H CPU, an RTX 3080 Ti GPU, and 64G of RAM.}.
To ensure reproducibility and foster future work in this direction, we open-source our implementation\footnote{<URL hidden for blind peer review>}.


\section{Experiments}\label{sec:experiments}

We use the normal-form matrices presented in Section~\ref{sec:egt} as our three tested interactions: \emph{Stag hunt} (SH), \emph{Hawk-Dove} (HD) and \emph{Rock-Paper-Scissors} (RPS).
Each individual agent has its own persistent learning rule: either PG (gradient descent on Equation~\ref{eq:pg11}) or LOLA (gradient descent on Equation~\ref{eq:fast_lola}).
The learning rate has no relevant impact on the dynamics presented in this paper other than varying their speed, thus we use unit learning rates everywhere.


\subsection{Scalability}
\begin{figure}
    \centering
    \begin{subfigure}[b]{0.32\textwidth}
        \centering
        \includegraphics[width=\textwidth, trim=0 0 0 0]{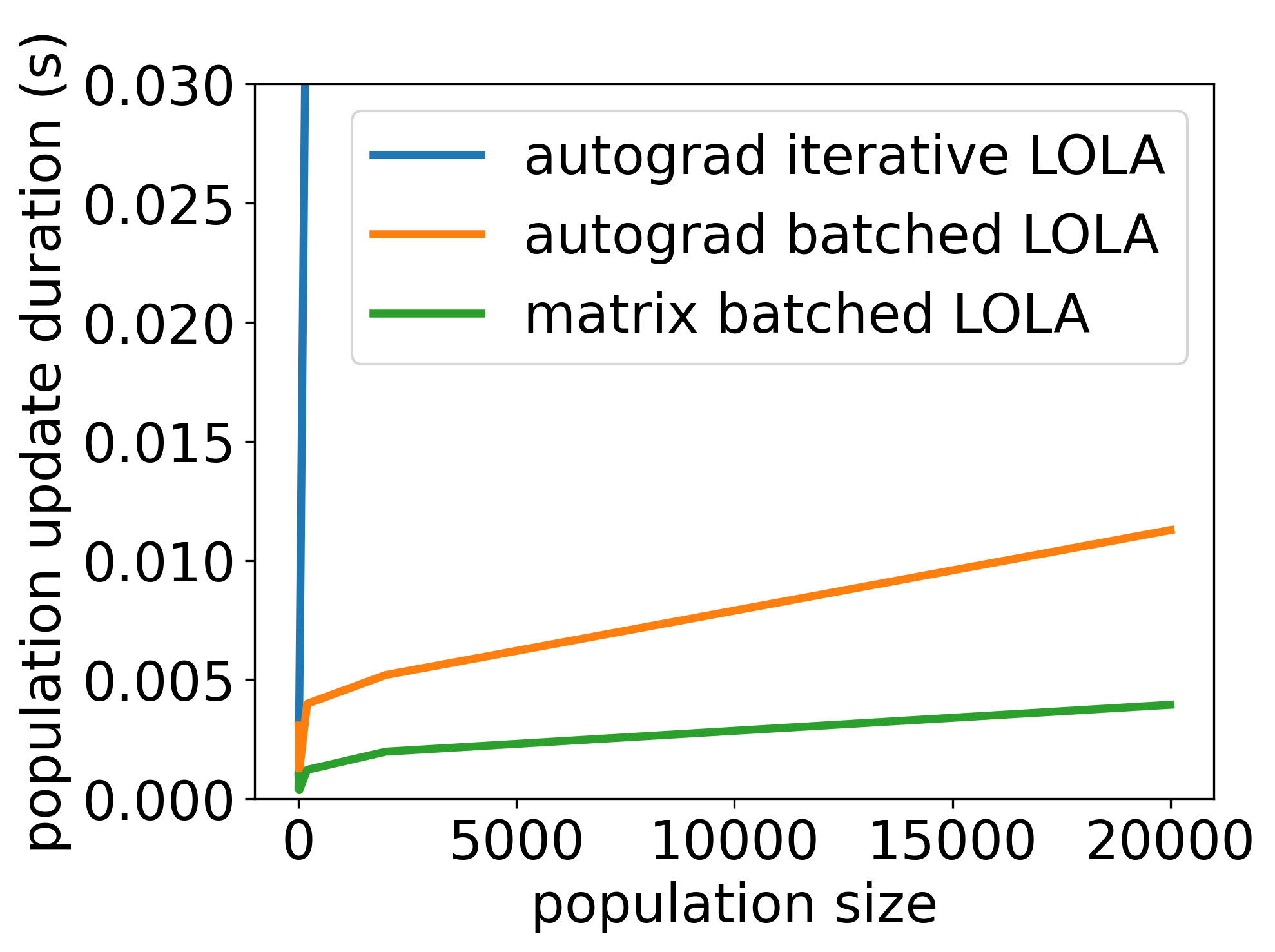}
        \caption{CPU}
    \end{subfigure}
    \hfill
    \begin{subfigure}[b]{0.32\textwidth}
        \centering
        \includegraphics[width=\textwidth, trim=0 0 0 0]{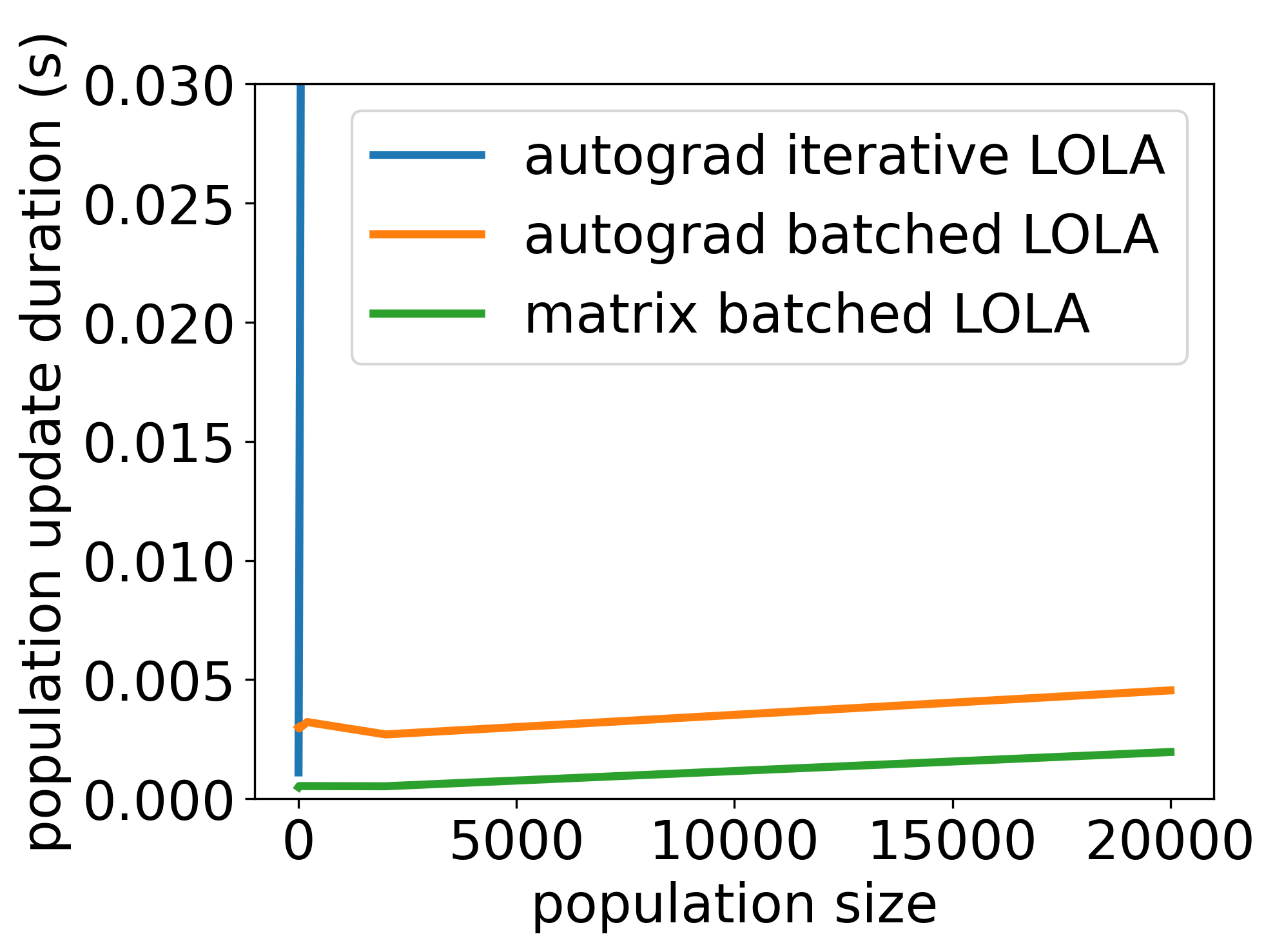}
        \caption{GPU}
    \end{subfigure}
    
    \caption{Duration of a full evolution step (lower is better)}
    \label{fig:timing}
\end{figure}

Figure~\ref{fig:timing} reports the computational performance of our approach (``matrix batched LOLA''), compared to two baselines.
The ``autograd iterative LOLA'' baseline reproduces how LOLA updates are usually performed in classical MARL scenarios: using PyTorch's autograd to compute the LOLA gradient, and updating the policies of all agents in an iterative fashion.
This baseline is clearly not a viable implementation for evolutionary-scale simulations and is only provided for illustration.
On the other hand, our ``autograd batched LOLA'' baseline is of more interest for future work.
While the ``matrix batched'' approach is faster, it is limited to single-shot multi-armed bandits\footnote{Section~\ref{sec:methods} is possible because the value function has a straightforward formulation in 2-agent multi-armed bandits.}.
In particular, the ``matrix batched'' approach does not allow episodic interactions.
Therefore, we have implemented the approach of Section~\ref{sec:batch} along with autograd, which yields a potentially more general implementation.
We present the performance of this alternative approach as ``autograd batched LOLA''.
Clearly, the main reason why we can perform these large-scale simulations at all is that the pairwise structure of EGT simulations enables batching interactions and policy updates.
Since our matrix-based implementation is faster for normal-form games, we use it in the following.

\begin{figure}
    \centering
    \begin{subfigure}[b]{0.32\textwidth}
        \centering
        \includegraphics[width=\textwidth, trim=0 0 10 0]{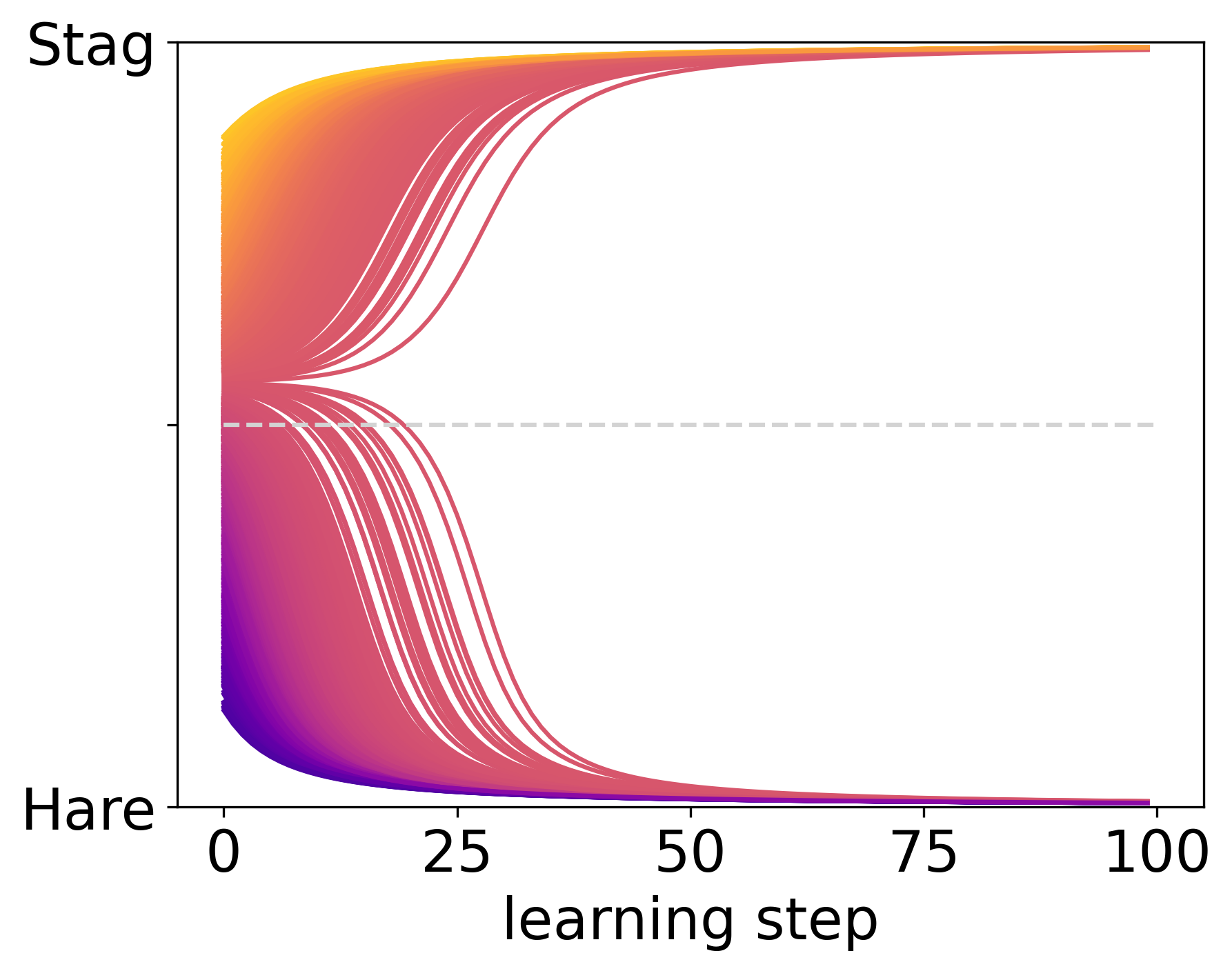}
        \caption{PG in SH ($s=1.8$)}
        \label{fig:sh_sp_pg}
    \end{subfigure}
    \hfill
    \begin{subfigure}[b]{0.32\textwidth}
        \centering
        \includegraphics[width=\textwidth, trim=0 0 10 0]{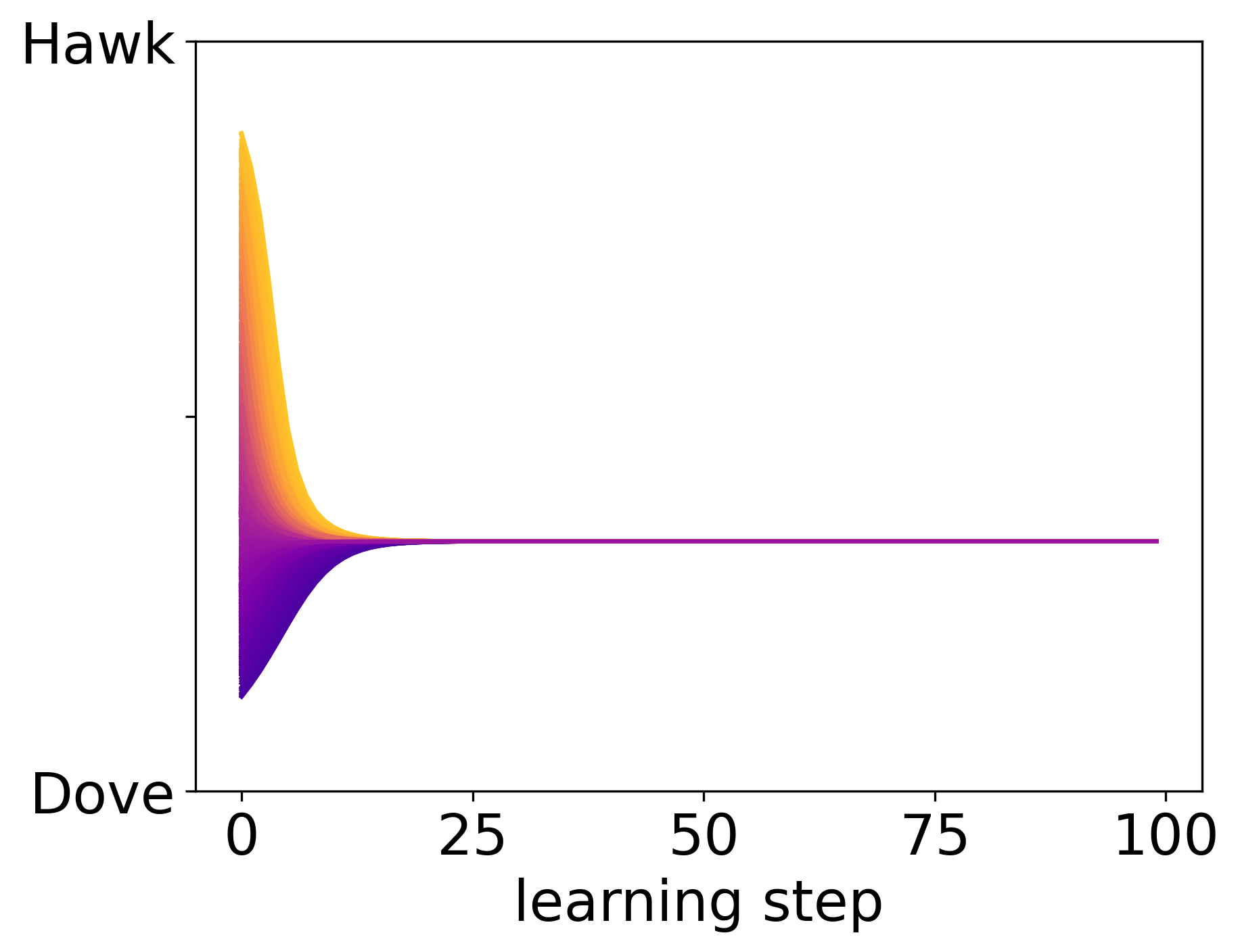}
        \caption{PG in HD ($f=-2$)}
        \label{fig:hd_sp_pg}
    \end{subfigure}
    \hfill
    \begin{subfigure}[b]{0.32\textwidth}
        \centering
        \includegraphics[width=\textwidth, trim=100 50 60 50, clip]{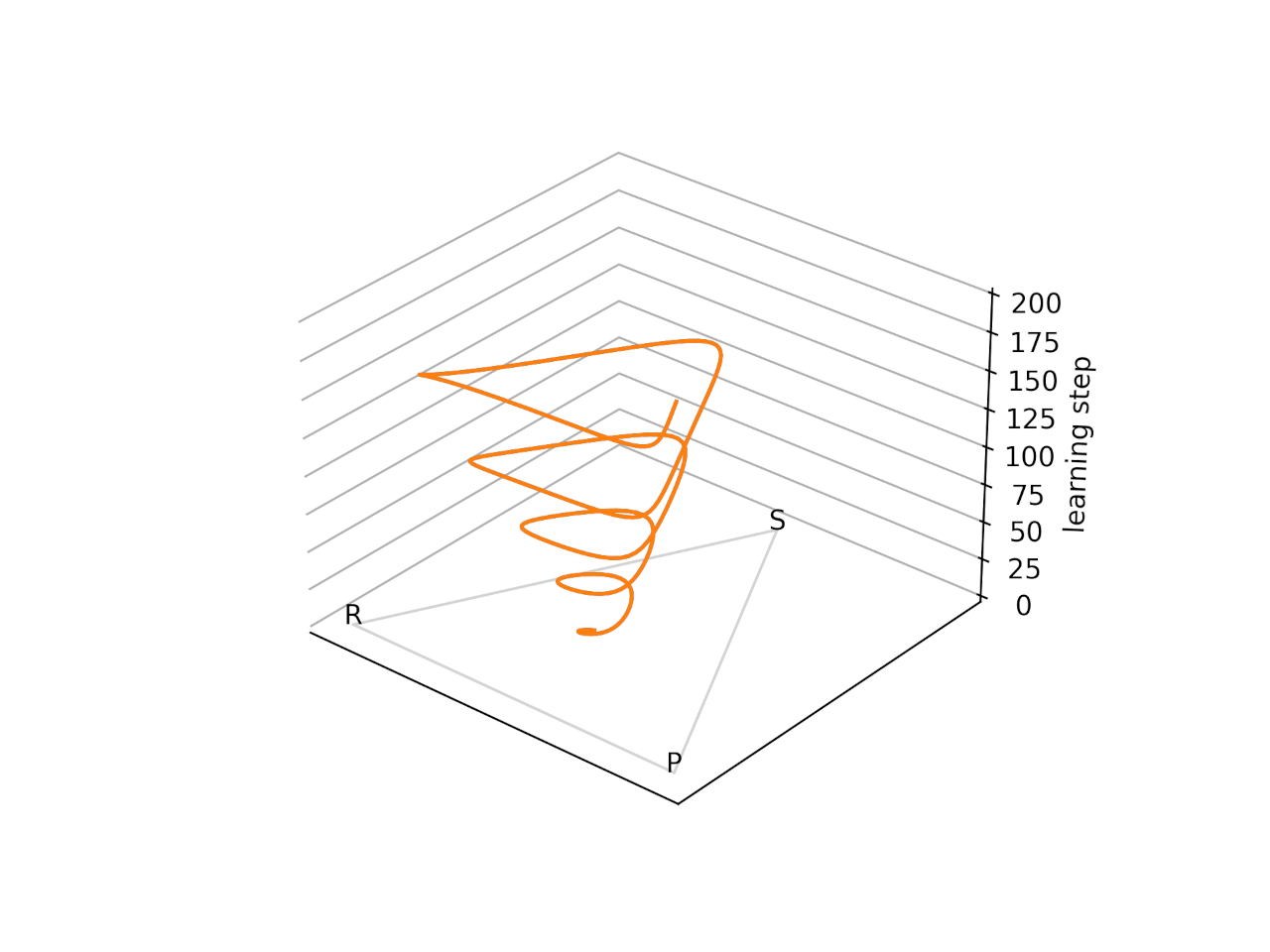}
        \caption{PG in RPS}
        \label{fig:rps_sp_pg}
    \end{subfigure}
    \hfill
    \begin{subfigure}[b]{0.32\textwidth}
        \centering
        \includegraphics[width=\textwidth, trim=10 0 0 0]{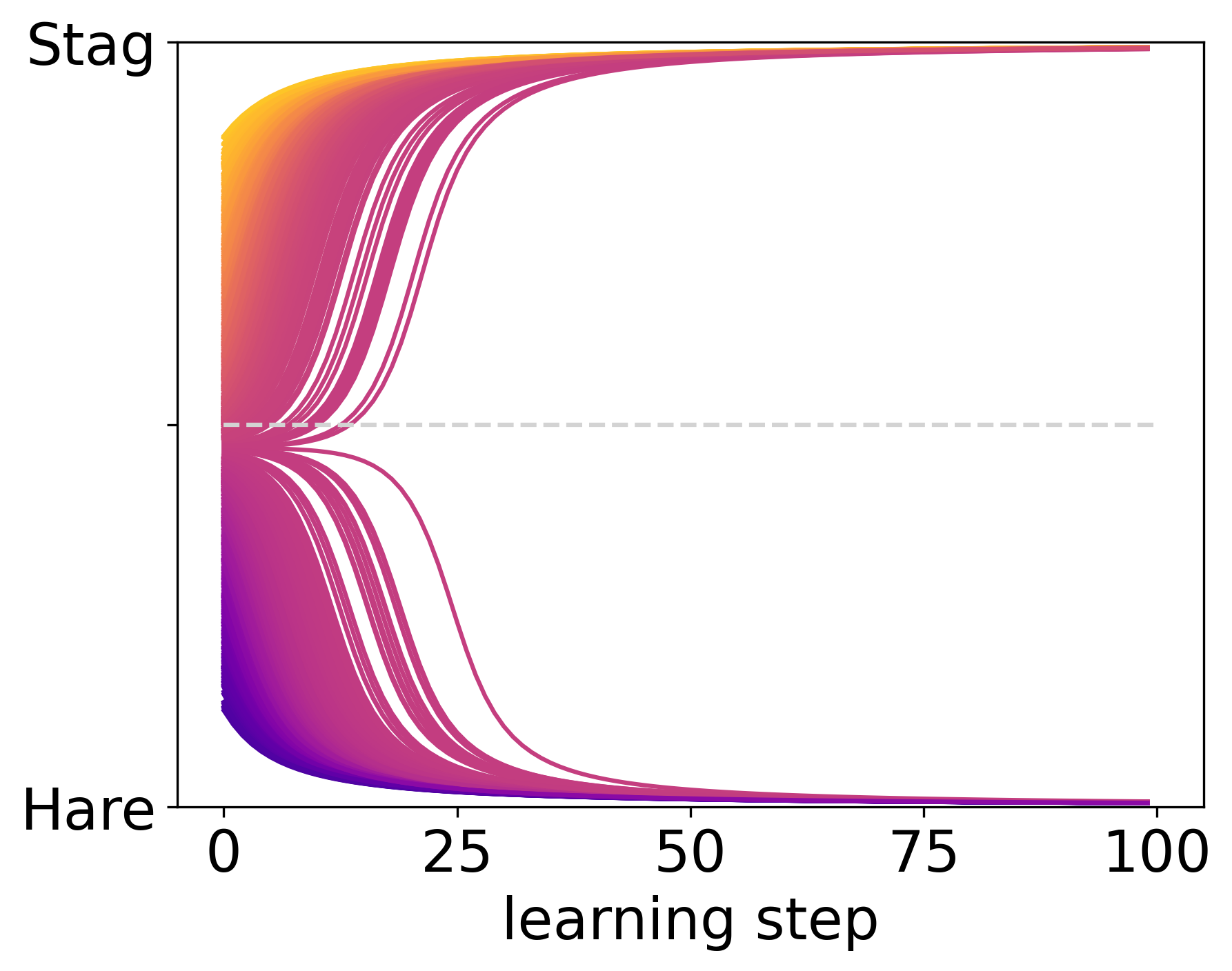}
        \caption{LOLA in SH ($s=1.8$)}
        \label{fig:sh_sp_lola}
    \end{subfigure}
    \hfill
    \begin{subfigure}[b]{0.32\textwidth}
        \centering
        \includegraphics[width=\textwidth, trim=10 0 0 0]{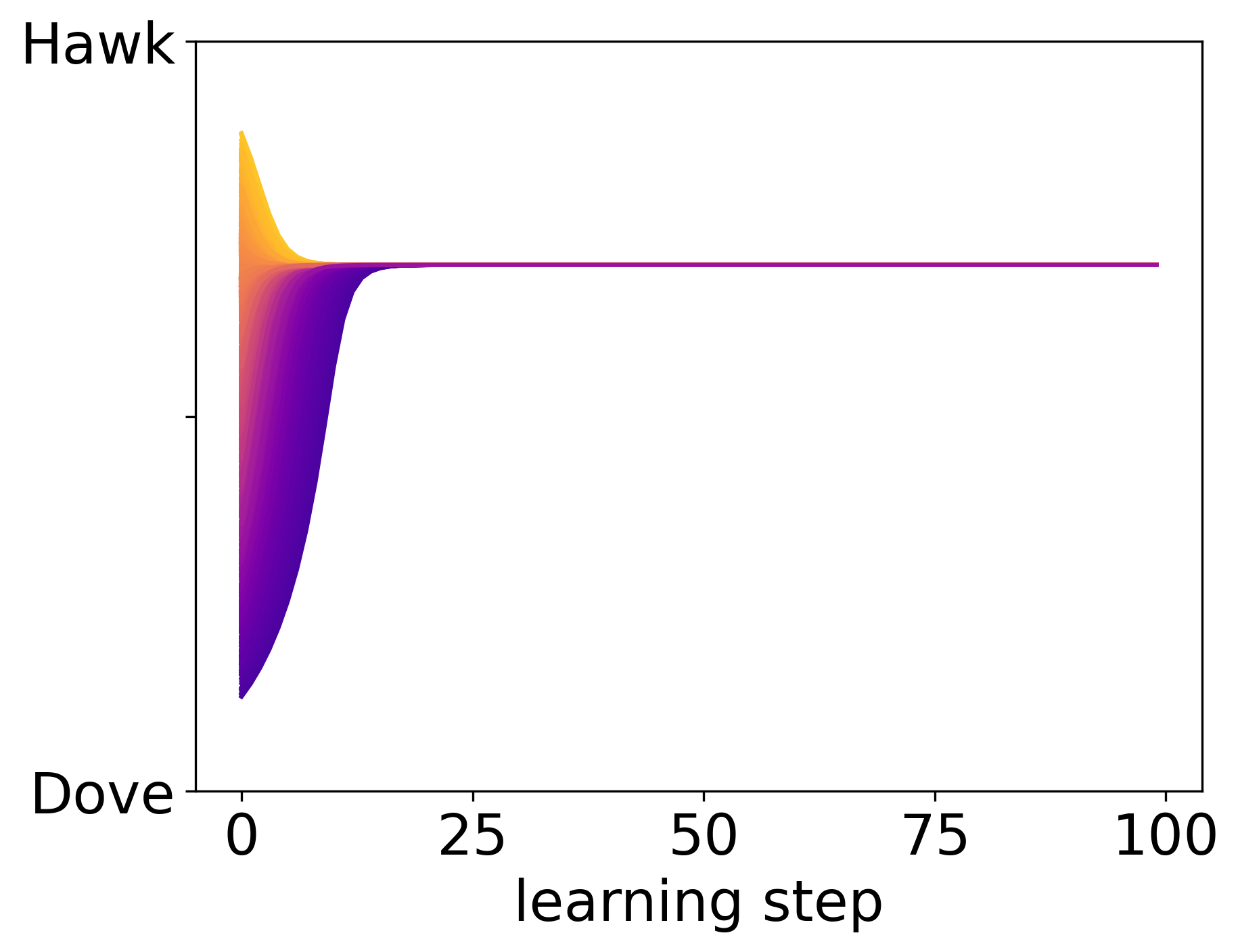}
        \caption{LOLA in HD ($f=-2$)}
        \label{fig:hd_sp_lola}
    \end{subfigure}
    \hfill
    \begin{subfigure}[b]{0.32\textwidth}
        \centering
        \includegraphics[width=\textwidth, trim=100 50 60 50, clip]{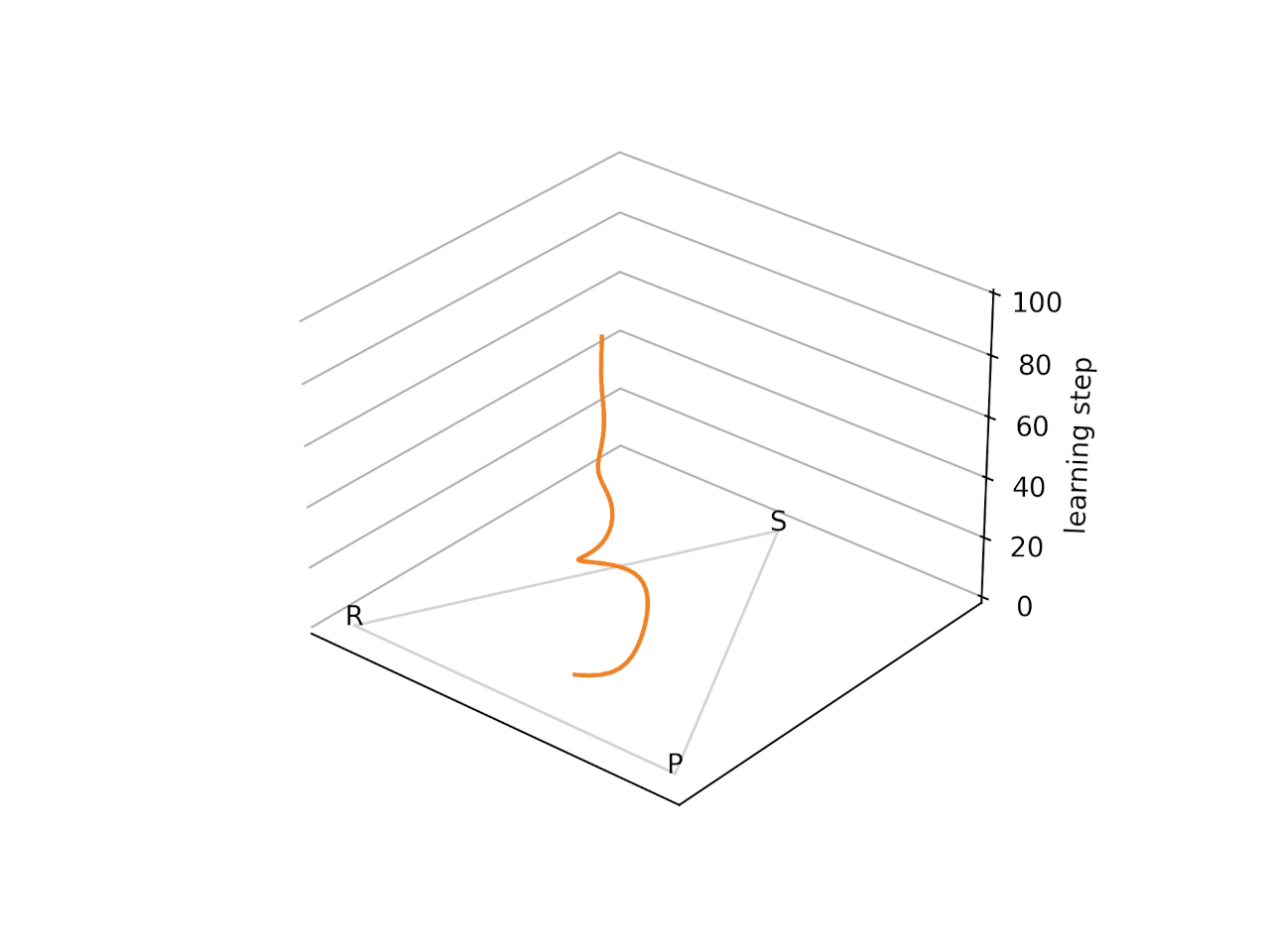}
        \caption{LOLA in RPS}
        \label{fig:rps_sp_lola}
    \end{subfigure}
    
    \caption{Self-play. In SH and HD, the color marks the initial policy.}
    \label{fig:sp_all}
\end{figure}

\subsection{Expected results}\label{sec:expected}


The methodology proposed in Section~\ref{sec:methods} is limited to very simple, single-shot bandit interactions between random pairs of learning agents.
Remember how, under the population-policy equivalence described in Section~\ref{sec:equivalence}, a population of pure-strategy agents can equivalently be seen as a stochastic policy over types.
In our setting, agents have full stochastic policies assigning non-zero probabilities to all available actions, but are still simple stateless multi-armed bandits.
Uniformly sampling a random pair of agents from such a population and then sampling from their policies is equivalent in expectation to sampling two actions from the average policy of all population members.
In other words, at least in expectation, we anticipate that the population dynamic will show some resemblance to self-play over the population's average policy (represented in Figure~\ref{fig:sp_all}).

\subsection{Empirical results}

\textbf{Stag Hunt.}
Figures~\ref{fig:sh_sp_pg} and \ref{fig:sh_sp_lola} show how a single self-play agent
learns against itself in Stag Hunt, via naive Policy Gradient and LOLA, respectively.
The vertical axis represents its policy, 
and the horizontal axis represents time expressed in learning steps.
Policies are color-coded by their initial configuration, with yellow policies starting close to the deterministic Stag policy and purple policies starting close to the deterministic Hare policy.
Notably, PG tends to converge to the individualistic Nash equilibrium (i.e., deterministic Hare) for most initial configurations, whereas LOLA tends to converge to the pro-social Nash equilibrium (i.e., deterministic Stag).
Notice that the forks are on different sides of the uniform random policy (middle tick), which is important because we will initialize all our population experiments with a Gaussian distribution around this neutral policy.
As explained in Section~\ref{sec:expected}, we expect the final population dynamic to follow a similar pattern.
Figures~\ref{fig:sh_evo_pg} and \ref{fig:sh_evo_lola} display the results of our first evolutionary simulation, featuring a population of 200,000 learning agents in Stag Hunt.
Dark shades of blue represent high concentrations of agents. 
An \emph{evolution step} corresponds to one learning step performed per agent in the population.
In Figure~\ref{fig:sh_evo_pg}, the population is exclusively formed of naive learners, and quickly converges to the individualistic policy, as predicted by Figure~\ref{fig:sh_sp_pg}.
In Figure~\ref{fig:sh_evo_lola}, the population is formed of only LOLA agents.
Contrary to the naive population, a population of non-stationarity aware learners such as LOLA evolves to unanimously adopt the superior pro-social equilibrium (i.e., deterministic Stag).
This effect is modulated by the payoff of the pro-social strategy, as measured in Figure~\ref{fig:sweep_sh}.
Appendix~\ref{ap:sh} further shows that, when enough opponent-learning aware learners are present in a mixed population, they become able to pull naive learners toward the pro-social strategy.

\textbf{Hawk-Dove}.
Figures~\ref{fig:hd_sp_pg} and \ref{fig:hd_sp_lola} show that, in Hawk-Dove, self-play converges to definite policies regardless of where training starts from.
Naive learning (Figure~\ref{fig:hd_sp_pg}) converges to the mixed Nash equilibrium. 
However, LOLA (Figure~\ref{fig:hd_sp_lola}) converges to another, inferior policy, where it selects Hawk 70\% of the time (which yields a smaller payoff for both players, and is not a Nash equilibrium).
A similar behavior has been described as ``arrogance'' in \cite{letcher2018stable}, where both LOLA learners make wrong assumptions about the response of their opponent and thus pull away from the equilibrium.
From these observations, one could imagine that all learning agents in the population would converge to these policies, similar to what we observed for SH, but this is not at all what happens in practice.
In HD, whether naive learning (Figure~\ref{fig:hd_evo_pg1}) or LOLA (Figure~\ref{fig:hd_evo_lola1}) is used as the learning rule of the entire population, it evolves into a mix of Hawks and Doves, most of them with close-to-deterministic policies, and in both cases with an average population policy that corresponds to the mixed Nash equilibrium. 
Notice that the convergence to deterministic strategies is however much slower than what we observed for SH, and it is in fact not clear whether this will eventually happen entirely, even after 10,000 steps (Appendix~\ref{ap:hd}).
Nonetheless, what can be observed from Figures~\ref{fig:hd_evo_pg1} and \ref{fig:hd_evo_pg2} is that LOLA learners converge faster to deterministic policies during early steps (shades of blue) but the population takes longer to stabilize (dotted black).
In additional experiments, we mixed LOLA and PG learners to see whether LOLA learners would be more inclined toward the Hawk strategy (as suggested by Figure~\ref{fig:hd_sp_lola}).
However, these experiments invalidated this hypothesis: 
half LOLA learners and half PG learners were present amongst both final sub-populations of Hawk-inclined and Dove-inclined individuals.

\textbf{Rock-Paper-Scissors.}
Our final population experiment takes place in the 3-action Rock-Paper-Scissors environment, used in EGT to explain the coexistence of competitively unbalanced species~\citep{allesina2011competitive}.
Interestingly, we will see that, while populations of naive learners agree with this explanation of sustainable diversity, populations of LOLA learners yield quite the opposite result.
Similarly to previous Sections, Figures~\ref{fig:rps_sp_pg} and \ref{fig:rps_sp_lola} show how 2-agent self-play behaves for both PG and LOLA.
Only 1 policy is displayed for readability (other initial conditions yield similar effects).
The triangle on the bottom of each plot represents the policy, and the vertical axis represents the number of learning steps.
It can be seen that PG slowly spirals outward from the mixed Nash equilibrium (due to performing straight policy updates with a non-zero learning rate following a circular vector field), whereas LOLA quickly spirals inward until it reaches the mixed Nash equilibrium.
Figures~\ref{fig:rps_evo_pg} and \ref{fig:rps_evo_lola} present the results of our evolutionary simulations in the RPS game.
The color-code follows the same principle as our previous population plots, with the bottom triangle being the policy, and evolution steps being the vertical dimension.
Similarly to Figure~\ref{fig:hd_evo_pg1}, Figure~\ref{fig:rps_evo_pg} shows that populations of naive learners evolve into 3 equally distributed groups of close-to-deterministic agents always playing Rock, Paper and Scissors respectively.
The reason why this happens is of interest, and is more clearly understood from Figure~\ref{fig:rps_evo_pg_triangle} in Appendix~\ref{ap:rps}.
In short, this dynamic results from the loss of plasticity modeled by the softmax function\footnote{A model very similar to the ``cost of motion'' described by \cite{mertikopoulos2018riemannian}.}.
At the beginning of evolution, naive learners are erratically moving around as they encounter all types of strategies.
However, after some time, agents get ``trapped'' near the border of the policy simplex, where gradients toward the opposite action are near-zero.
After a long time, three groups of near-deterministic agents emerge, and a small number of them continuously escape toward the strategy that counters the majority group, which eventually creates a new majority, and so on, yielding a cyclic evolution pattern.
In other words, diversity emerges from populations of naive learners in the RPS model.
On the other hand, Figure~\ref{fig:rps_evo_lola} tells the opposite story about populations of LOLA agents, which instead quickly and unanimously converge to the mixed Nash equilibrium of this game.


\subsection{Remark}
From our simulation results, it looks like the mean policy averaged over the entire population always converges near a Nash equilibrium of the game, even when the learning rule itself does not converge to this equilibrium in the conventional 2-agent MARL setting (see Figures~\ref{fig:hd_sp_lola} and \ref{fig:rps_sp_pg}).
This property is however merely a consequence of the uniform random opponent matching scheme that we chose to implement in this paper.
For instance, let us consider an extreme opposite scheme, where all pairs would instead interact persistently.
All individual pairs would then converge to the pure Nash equilibrium in the Hawk-Dove game (that is, exactly one deterministic Hawk and one deterministic Dove per pair): this would average to a uniform random policy, as opposed to Figures~\ref{fig:hd_evo_pg1} and \ref{fig:hd_evo_lola1}.
In reality, partner selection is more structured~\citep{ana20partner} and can lead to different outcomes, which we plan to explore in the future.




\section{Conclusions}
We have presented a methodology enabling large-scale evolutionary simulations of independent learning agents, for both naive (Policy Gradient) and advanced non-stationarity-aware (LOLA) learning rules. 
We have demonstrated the scalability of our approach by performing very-large-scale evolutionary simulations of 200,000 independent learning agents, interacting in the classic games of Stag Hunt, Hawk-Dove and Rock-Paper-Scissors.
Our work essentially explores the effect of Multi-Agent Reinforcement Learning on the usual model of evolution adopted in Evolutionary Game Theory, and demonstrates compelling social dynamics originating from both naive and non-stationarity-aware learners.
While the approach presented in this paper is specifically designed for normal-form matrix games (i.e., stateless 2-player multi-armed bandits), exploring stateful episodic scenarios with a gradient-based approach might be possible, and is an avenue for future work.

\appendix







\bibliography{main}
\bibliographystyle{rlj}

\beginSupplementaryMaterials

\section{Related literature}\label{ap:related}
This paper is a first step toward describing the fast-paced, social evolution that stems from individuals actively optimizing their own fitness in real-world societies and economies.
While the idea is probably not novel~\citep{tuyls2005evolutionary}, no scientific progress has been made in developing a theory around it so far.
In fact, Evolutionary Game Theory (EGT) usually avoids MARL entirely, as it introduces a substantial amount of theoretical complexity when modeling population dynamics and does not relate to the stochasticity-driven model of genetic evolution accepted so far (a model however recently challenged by the work of \cite{beavan2024contingency}).
Before our work, MARL was complex to simulate at evolutionary scale and very few papers have attempted anything similar.
Thus, the aim of this Section is to motivate our line of work by providing a brief overview of the most closely related fields of research and positioning ourselves with respect to their literature.

\textbf{Evolutionary Game Theory.}
Originally rooted in biology, EGT extends classical Game Theory to the study of evolving populations.
\cite{smith1973logic} laid its mathematical foundations and introduced the concept of Evolutionarily Stable Strategies, formalizing how different types are maintained in genetic evolution. 
\cite{axelrod1981evolution} famously described how cooperation naturally arises in the framework of EGT. 
More recently, \cite{nowak2006five} proposed five mechanisms fostering the evolution of cooperation: kin selection, direct reciprocity, indirect reciprocity, network reciprocity, and - an idea that has a long history of controversy amongst evolutionary biologists - group selection.
Finally, \cite{mertikopoulos2018riemannian} extended the mathematical framework available for studying EGT, by formalizing population dynamics (such as the Replicator Dynamic) under the more general class of Riemannian game dynamics. 

\textbf{Opinion dynamics.}
In parallel to its original motivation in biology, EGT has gathered interest from different fields, especially economy and social dynamics.
In an influential book promoting the ``selfish'' gene-centered view of evolution, the ethologist \cite{dawkins1976selfish} speculated about ``memes'', an alleged generalization of genetics to cultural dynamics.
Amongst other assumptions, his idea was based on the belief that \emph{replication of the fittest}, the gradient-free revision protocol through which populations of non-learning agents are thought to evolve in nature~\citep{apesteguia2007imitation}, extends beyond biology and, in particular, to opinions and strategies.
Because this belief seems relevant to situations where individuals learn via imitation, it has later inspired a large body of work under the name ``imitation dynamics'' \citep{sandholm2010population, xia2011opinion}.
Mentioning this competing line of thought is interesting, as our work studies the consequences of fundamentally different assumptions.
Namely, whereas genetic evolution arguably stems from replication dynamics and random mutations (which may also be the case for social strategies/opinions to some extent), the dynamics that we specifically study in this paper instead stem from continual learning.
We do not consider any replication process, but only a gradient-informed, learning-based ``mutation'' process.

\textbf{Population-Based Training} (PBT)~\citep{jaderberg2017population} is a line of efficient bio-inspired MARL approaches illustrating the potential complementarity of both views. 
In PBT, two processes coexist:
an inner training loop lets a group of agents learn via MARL rules, while an outer genetic loop selects and replicates the fittest few resulting policies.

\textbf{Opponent shaping.}
Advanced MARL rules are able to take advantage of the non-stationarity stemming from the learning processes of other agents.
For instance, an agent can shape its opponent by learning to share its own rewards~\citep{lupu19gifting,yang20lio}.
However, this involves learning to reward opponents when they exhibit favorable behaviors, which exacerbates non-stationarity as credit assignment gets harder.
Opponent-learning awareness methods such as LOLA overcome this issue by instead differentiating through the learning step of other agents~\citep{foerster2017learning}.
An extensive line of research recently emerged in this direction, attempting to fix diverse issues in the original LOLA formulation~\citep{foerster18adice, letcher2018stable, Lu2022ModelFreeOS, Willi2022COLACL}. 
In this paper, we find a way to simulate LOLA in large populations.

\textbf{Simulation.}
Due to the high theoretical complexity of MARL-based population dynamics, our paper focuses on empirical simulation.
Even when using simple replication-based revision protocols, resorting to simulation is common in EGT.
For example, \cite{santos2005scale} famously simulated how structured locality fosters cooperation.
\cite{julia18comp} confirmed various theoretical predictions regarding cycles and their effect in games by simulating the evolution of non-learning strategies in iterated prisoner's dilemmas.
Interestingly, EGT simulations have also been used by \cite{omi2019alpha}, who proposed to simulate imitation of the fittest to rank policies within a population of readily-trained agents.
Finally, one work of particular relevance to ours has been conducted by \cite{yang2020mfmarl}, who used mean-field theory in a classical MARL scenario to reduce the environment dimensionality.
In their proposed approach, naive learners approximate neighboring agents as one single, "mean-field" opponent.
This essentially transforms $n$-agent MARL into pairwise MARL, thus reducing the environment complexity from the point of view of individual agents.

\textbf{Parallels between EGT and single-agent RL.}
The population-policy equivalence described by \cite{bloembergen2015evolutionary} yields interesting parallels between single-agent Reinforcement Learning and the Replicator Dynamic.
In particular, the resemblance of Policy Gradient with the Replicator Dynamic noted in Section~\ref{sec:analytical_pg} was further studied by \cite{hennes2020neural}.
From this observation, they derived a single-agent algorithm that bypasses the loss of plasticity introduced by the softmax architecture of Equation~\ref{eq:model}.
However, beyond the fact that their line of work uses concepts from RL and EGT, it is mostly unrelated to ours and we cite it here to clear a confusion made by early readers of our work: they are interested in finding high-performance single-agent RL algorithms, whereas we are interested in characterizing the population dynamics that stem from MARL revision protocols when individuals actively learn in massively multi-agent, evolutionary settings.




\section{Policy gradient}\label{ap:pg}

We derive an analytical formulation of the PG update in symmetric normal-form games:

\begin{align*}
    \nabla_{\theta_1} v_1 &= \nabla_{\theta_1} P_1^\top A P_2 \\
    &= (\mbox{diag}(P_1) -P_1 P_1^\top) A P_2\notag \\
    &= \mbox{diag}(P_1) A P_2 - P_1 v_1 \notag \\
    &= P_1 \odot (A P_2 - v_1 \boldsymbol{1})\notag \\
    &= P_1 \odot (A P_2 - \boldsymbol{1} P_1^\top A P_2)\notag \\
    &= P_1 \odot (Q_1 - \boldsymbol{1} v_1)
\end{align*}

\section{LOLA}\label{ap:lola}

We now derive an analytical formulation of the LOLA update in symmetric normal-form games, similar to what we found for PG in Section~\ref{sec:analytical_pg}.
We are missing three terms from Equation~\ref{eq:lola}:
\begin{itemize}
    \item $\nabla_{\theta_2} v_1$
    \item $\nabla_{\theta_1} \nabla_{\theta_2} v_1$
    \item $\nabla_{\theta_1} \nabla_{\theta_2} v_2$
\end{itemize}

To compute the first term, we note that $v_1 =  P_1^\top A P_2$ is a scalar and thus can also be written $v_1 =  P_2^\top A^\top P_1$.
We can then compute this term similarly to PG:
\begin{align}
    \nabla_{\theta_2} v_1 &= \nabla_{\theta_2} P_2^\top A^\top P_1 \notag \\
    &= (\mbox{diag}(P_2) - P_2 P_2^\top) A^\top P_1\notag \\
    &= P_2 \odot (A^\top P_1 - v_1 \boldsymbol{1})\notag \\
    &= P_2 \odot (A^\top P_1 - \boldsymbol{1} P_2^\top A^\top P_1)\notag \\
    &= P_2 \odot (I - \boldsymbol{1} P_2^\top) A^\top P_1 \label{ea:lola21}
\end{align}

Computing the two remaining terms is also possible. \\
Let us start with $\nabla_{\theta_1} \nabla_{\theta_2} v_2$:
\begin{align}
    \nabla_{\theta_1} \nabla_{\theta_2} v_2 &= \nabla_{\theta_1} \nabla_{\theta_2} P_2^\top A P_1 \label{eq:ref1} \\
    &= \nabla_{\theta_1} (\mbox{diag}(P_2) - P_2 P_2^\top) A P_1\notag \\
    &= (\mbox{diag}(P_2) - P_2 P_2^\top) A (\mbox{diag}(P_1) - P_1 P_1^\top)
\end{align}
While it would already possible to implement this formulation, we further derive a more efficient implementation in Appendix~\ref{ap:fast_lola}:
\begin{equation} \label{eq:lola122}
    \nabla_{\theta_1} \nabla_{\theta_2} v_2 = T \odot (I - \boldsymbol{1} P_2^\top) A (I - P_1 \boldsymbol{1}^\top)
\end{equation}
where $T:=P_2 P_1^\top$. 

Computing $\nabla_{\theta_1} \nabla_{\theta_2} v_1$ is fairly straightforward:
\begin{align}
    \nabla_{\theta_1} \nabla_{\theta_2} v_1
    &= \nabla_{\theta_1} \nabla_{\theta_2} P_2^\top A^\top P_1 \notag\\
    &= \nabla_{\theta_1} \nabla_{\theta_2} P_2^\top B P_1 & (B:=A^\top)\notag \\
    &= T \odot (I - \boldsymbol{1} P_2^\top) B (I - P_1 \boldsymbol{1}^\top) & (\text{c.f. \ref{eq:ref1},\ref{eq:lola122}}) \\
    &= T \odot (I - \boldsymbol{1} P_2^\top) A^\top (I - P_1 \boldsymbol{1}^\top) \label{eq:lola121}
\end{align}

Substituting Equations~\ref{eq:pg11}, \ref{eq:pg22}, \ref{ea:lola21}, \ref{eq:lola122} and \ref{eq:lola121} in Equation~\ref{eq:lola} yields the following analytical formulation of the LOLA gradient in the symmetric normal-form game defined by matrix $A$:
\begin{align}
\nabla_{\theta_1} v_1(\theta_1, \theta_2 + \Delta \theta_2)
& \approx P_1 \odot X_1 A P_2 \nonumber\notag \\
& +\eta (T^\top \odot X_1 A X_2^\top) (P_2 \odot X_2 A P_1) \nonumber\notag\\
& +\eta (T^\top \odot X_1 A^\top X_2^\top) (P_2 \odot X_2 A^\top P_1)
\end{align}
where $X_1 := I - \boldsymbol{1} P_1^\top$, $X_2 := I - \boldsymbol{1} P_2^\top$, and $T:=P_2 P_1^\top$.

\section{Second-order policy gradients}\label{ap:fast_lola}

In this Section, we show that:

$$
\nabla_{\theta_1} \nabla_{\theta_2} v_2 = T \odot (I - \boldsymbol{1} P_2^\top) A (I - P_1 \boldsymbol{1}^\top)
$$
where $T:=P_2 P_1^\top$ is agent~$2$'s transition matrix.

\begin{proof}

\begin{align}
    \nabla_{\theta_1} \nabla_{\theta_2} v_2 &= \mbox{diag}(P_2) - P_2 P_2^\top) A (\mbox{diag}(P_1) - P_1 P_1^\top) \nonumber \\
    &= \mbox{diag}(P_2) A~\mbox{diag}(P_1)
-\mbox{diag}(P_2) A P_1 P_1^\top
-P_2 P_2^\top A~\mbox{diag}(P_1)
+P_2 P_2^\top A P_1 P_1^\top \label{eq:ap1}
\end{align}

\noindent
Note that, for $X, Y \in \mathbb{R}^n$:
$$
X Y^\top = X \boldsymbol{1}^\top \odot \boldsymbol{1} Y^\top
$$
since:
$$
\begin{pmatrix}
x_{1} y_{1} & x_{1}y_{2} & \dots & x_{1}y_{n}\\
x_{2}y_{1} & x_{2} y_{2} & \dots & x_{2}y_{n}\\
\vdots & \vdots &  & \vdots\\
x_{n}y_{1} & x_{n}y_{2} & \dots & x_{n}y_{n}
\end{pmatrix}
= \begin{pmatrix}
x_{1} & x_{1} & \dots & x_{1}\\
x_{2} & x_{2} & \dots & x_{2}\\
\vdots & \vdots &  & \vdots\\
x_{n} & x_{n} & \dots & x_{n}
\end{pmatrix}
\odot \begin{pmatrix}
y_{1} & y_{2} & \dots & y_{n}\\
y_{1} & y_{2} & \dots & y_{n}\\
\vdots & \vdots &  & \vdots\\
y_{1} & y_{2} & \dots & y_{n}
\end{pmatrix}
$$
Also, for $M\in \mathbb{R}^{n,n}$ note that:
$$
\text{diag}(X) M = X \boldsymbol{1}^\top \odot M
$$
since:
$$
\begin{pmatrix}
x_{1} m_{1,1} & x_{1}m_{1,2} & \dots & x_{1}m_{1,n}\\
x_{2}m_{2,1} & x_{2}m_{2,2} & \dots & x_{2}m_{2,n}\\
\vdots & \vdots &  & \vdots\\
x_{n}m_{n,1} & x_{n}m_{n,2} & \dots & x_{n}m_{n,n}
\end{pmatrix}
= \begin{pmatrix}
x_{1} & x_{1} & \dots & x_{1}\\
x_{2} & x_{2} & \dots & x_{2}\\
\vdots & \vdots &  & \vdots\\
x_{n} & x_{n} & \dots & x_{n}
\end{pmatrix}
\odot \begin{pmatrix}
m_{1,1} & m_{1,2} & \dots & m_{1,n}\\
m_{2,1} & m_{2,2} & \dots & m_{2,n}\\
\vdots & \vdots &  & \vdots\\
m_{n,1} & m_{n,2} & \dots & m_{n,n}
\end{pmatrix}
$$
And similarly:
$$
M \text{diag}(X)
= (\text{diag}(X) M ^\top)^\top
= (X \boldsymbol{1}^\top \odot M^\top)^\top
= M \odot \boldsymbol{1} X^\top
$$
\newline
\noindent
So, taking a closer look at each term in Equation~\ref{eq:ap1}:

\begin{align*}
    \mbox{diag}(P_2) A~\mbox{diag}(P_1) &= T \odot A
\end{align*}
\begin{align*}
    \mbox{diag}(P_2) A P_1 P_1^\top &= P_2 \boldsymbol{1}^\top \odot A P_1 P_1^\top \\
    &= P_2 \boldsymbol{1}^\top \odot A P_1 \boldsymbol{1}^\top \odot \boldsymbol{1} P_1^\top \\
    &= P_2 \boldsymbol{1}^\top \odot \boldsymbol{1} P_1^\top \odot A P_1 \boldsymbol{1}^\top \\
    &= T \odot A P_1 \boldsymbol{1}^\top
\end{align*}
\begin{align*}
    P_2 P_2^\top A~\mbox{diag}(P_1) &= P_2 P_2^\top A \odot \boldsymbol{1} P_1^\top \\
    &= P_2 \boldsymbol{1}^\top \odot \boldsymbol{1} P_2^\top A \odot \boldsymbol{1} P_1^\top \\
    &= P_2 \boldsymbol{1}^\top \odot \boldsymbol{1} P_1^\top \odot \boldsymbol{1} P_2^\top A \\
    &= T \odot \boldsymbol{1} P_2^\top A
\end{align*}
\begin{align*}
    P_2 P_2^\top A P_1 P_1^\top = P_2 v_2 P_1^\top &= T \odot v_2 \boldsymbol{1} \boldsymbol{1}^\top
\end{align*}
\newline
\noindent
This enables writing the LOLA second-order gradient as:
$$
\nabla_{\theta_1} \nabla_{\theta_2} v_2 =
T \odot
(
A
-A P_1 \boldsymbol{1}^\top
-\boldsymbol{1} P_2^\top A
+v_2 \boldsymbol{1} \boldsymbol{1}^\top
)
$$
\newline







\noindent
The term between parentheses can be factorized:
\begin{align*}
    A-A P_1 \boldsymbol{1}^\top - \boldsymbol{1} P_2^\top A + v_2 \boldsymbol{1} \boldsymbol{1}^\top
    &=A-A P_1 \boldsymbol{1}^\top - \boldsymbol{1} P_2^\top A + \boldsymbol{1} v_2 \boldsymbol{1}^\top \\
    &=A-A P_1 \boldsymbol{1}^\top - \boldsymbol{1} P_2^\top A + \boldsymbol{1} P_2^\top A P_1 \boldsymbol{1}^\top \\
    &= (I - \boldsymbol{1} P_2^\top)A - (I - \boldsymbol{1} P_2^\top) A P_1 \boldsymbol{1}^\top \\
    &= (I - \boldsymbol{1} P_2^\top)(A - A P_1 \boldsymbol{1}^\top) \\
    &= (I - \boldsymbol{1} P_2^\top) A (I - P_1 \boldsymbol{1}^\top)
\end{align*}

So:

$$
\nabla_{\theta_1} \nabla_{\theta_2} v_2 = T \odot (I - \boldsymbol{1} P_2^\top) A (I - P_1 \boldsymbol{1}^\top)
$$

\end{proof}

\section{Stag Hunt}\label{ap:sh}

\begin{figure}[h]
    \centering
    \begin{subfigure}[b]{0.45\textwidth}
        \centering
        \includegraphics[width=\textwidth, trim=0 0 0 0]{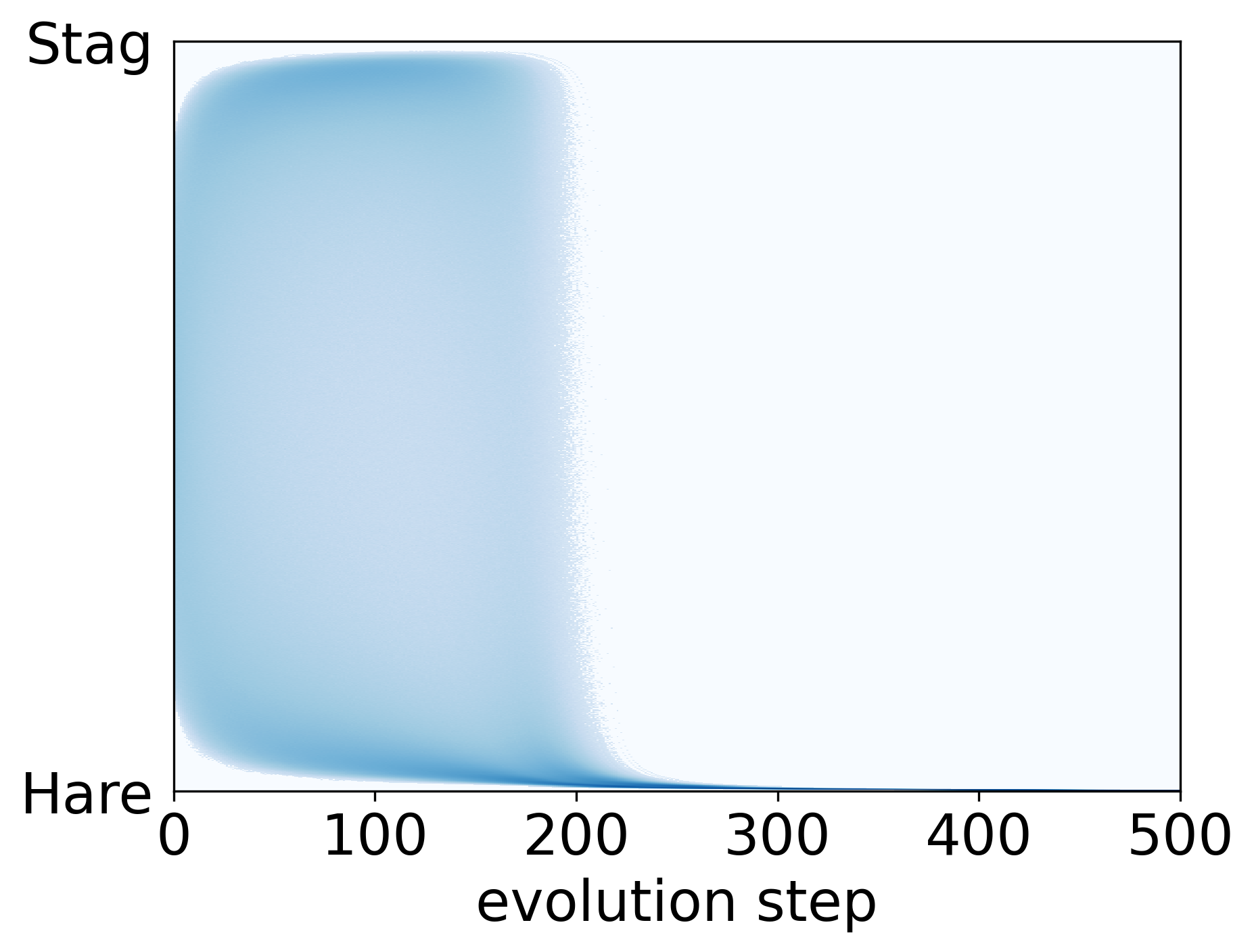}
        \caption{85\% LOLA}
        \label{fig:sh_evo_85}
    \end{subfigure}
    \hfill
    \begin{subfigure}[b]{0.45\textwidth}
        \centering
        \includegraphics[width=\textwidth, trim=0 0 0 0]{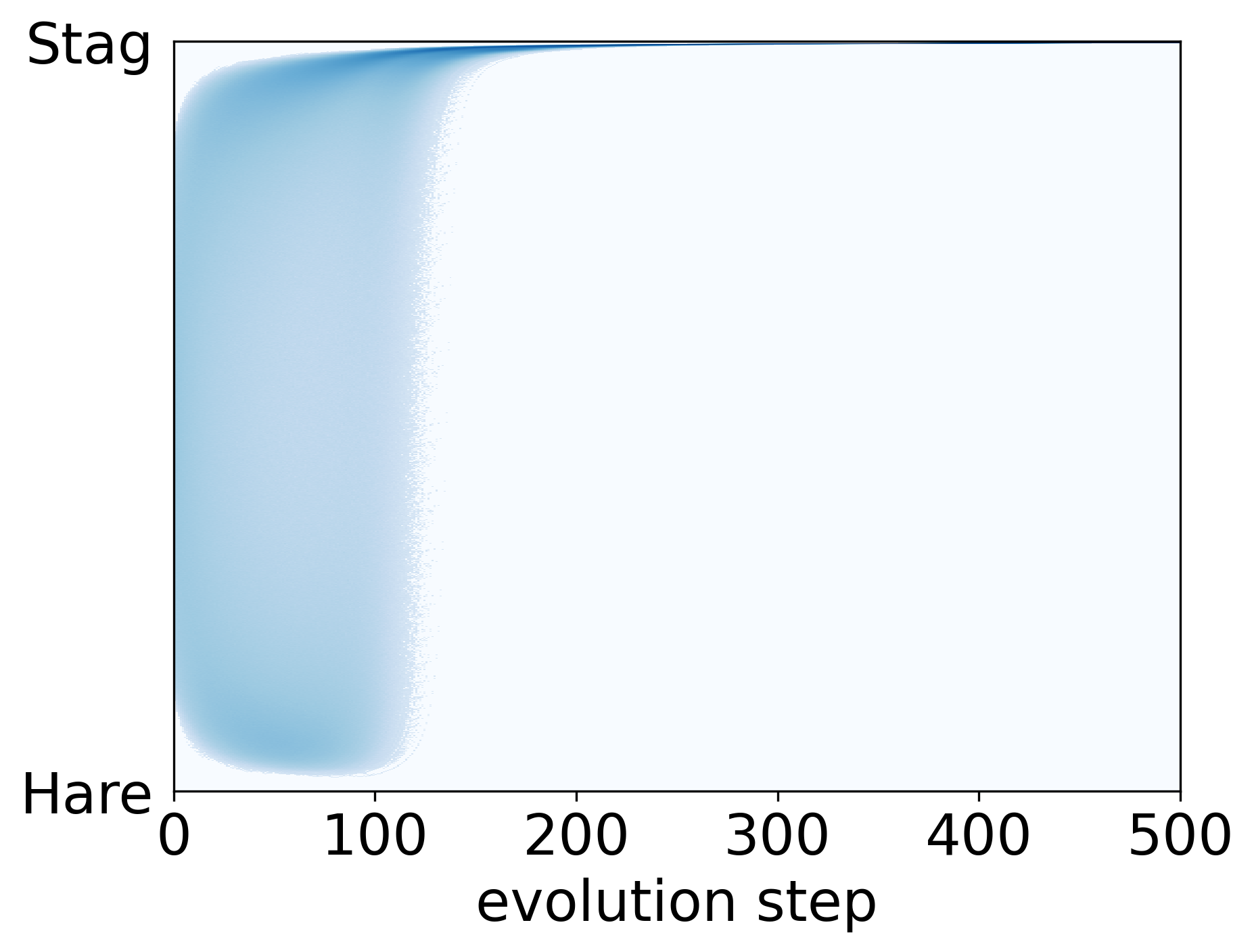}
        \caption{86\% LOLA}
        \label{fig:sh_evo_86}
    \end{subfigure}
    
    \caption{Mixed PG and LOLA in Stag Hunt ($s=1.8$). When more than 86\% of the population is made of LOLA agents, opponent-aware learners bring the entire population to the pro-social equilibrium (NB: the higher $s$ is, the lower this threshold becomes; it reaches 0\% when $s=2$).}
    \label{fig:sh_evo_ap}
\end{figure}

\section{Hawk Dove}\label{ap:hd}

\begin{figure}[h]
    \centering
    \begin{subfigure}[b]{0.45\textwidth}
        \centering
        \includegraphics[width=\textwidth, trim=0 0 0 0]{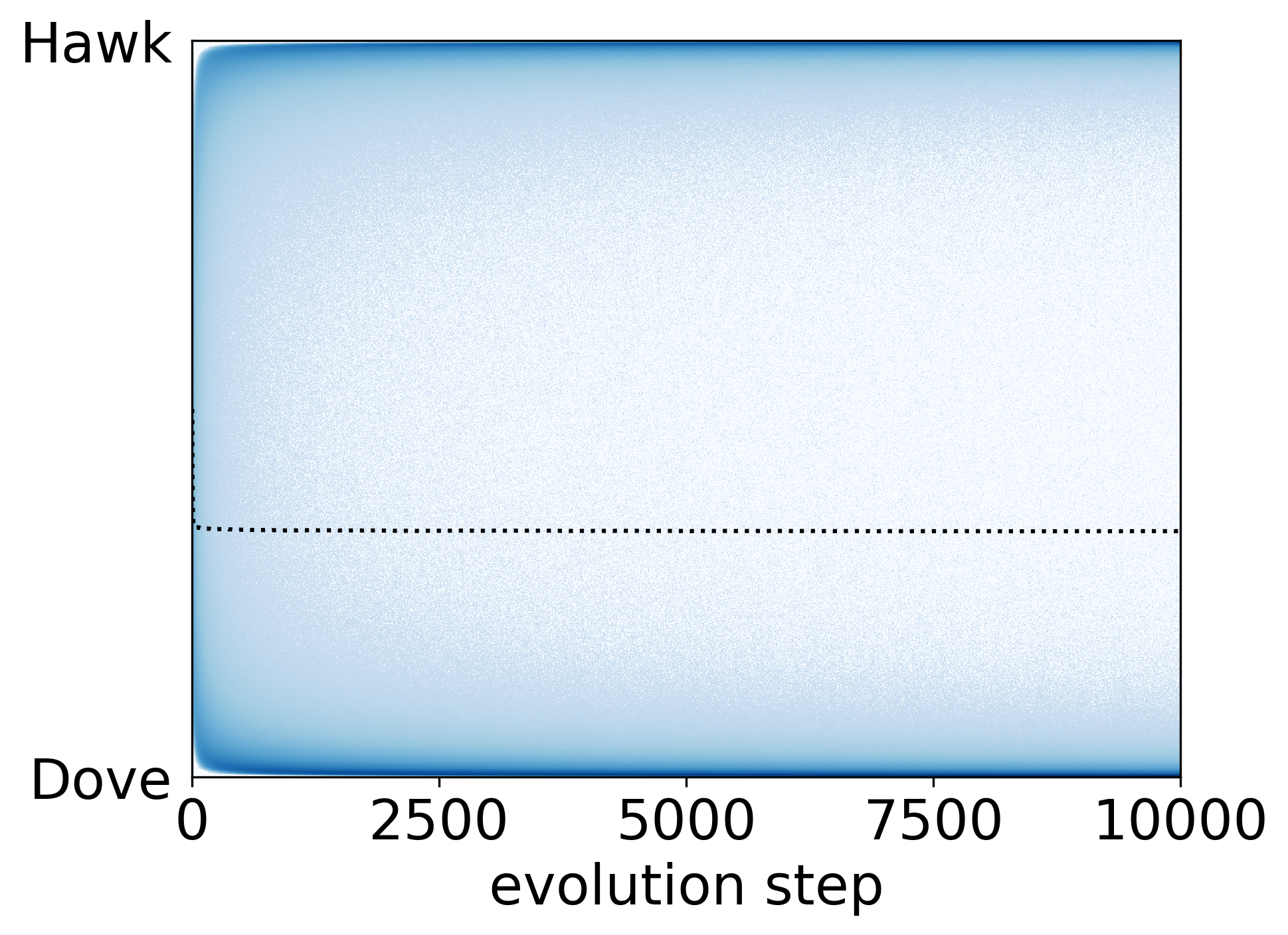}
        \caption{PG late evolution}
        \label{fig:hd_evo_pg2}
    \end{subfigure}
    \hfill
    \begin{subfigure}[b]{0.45\textwidth}
        \centering
        \includegraphics[width=\textwidth, trim=0 0 0 0]{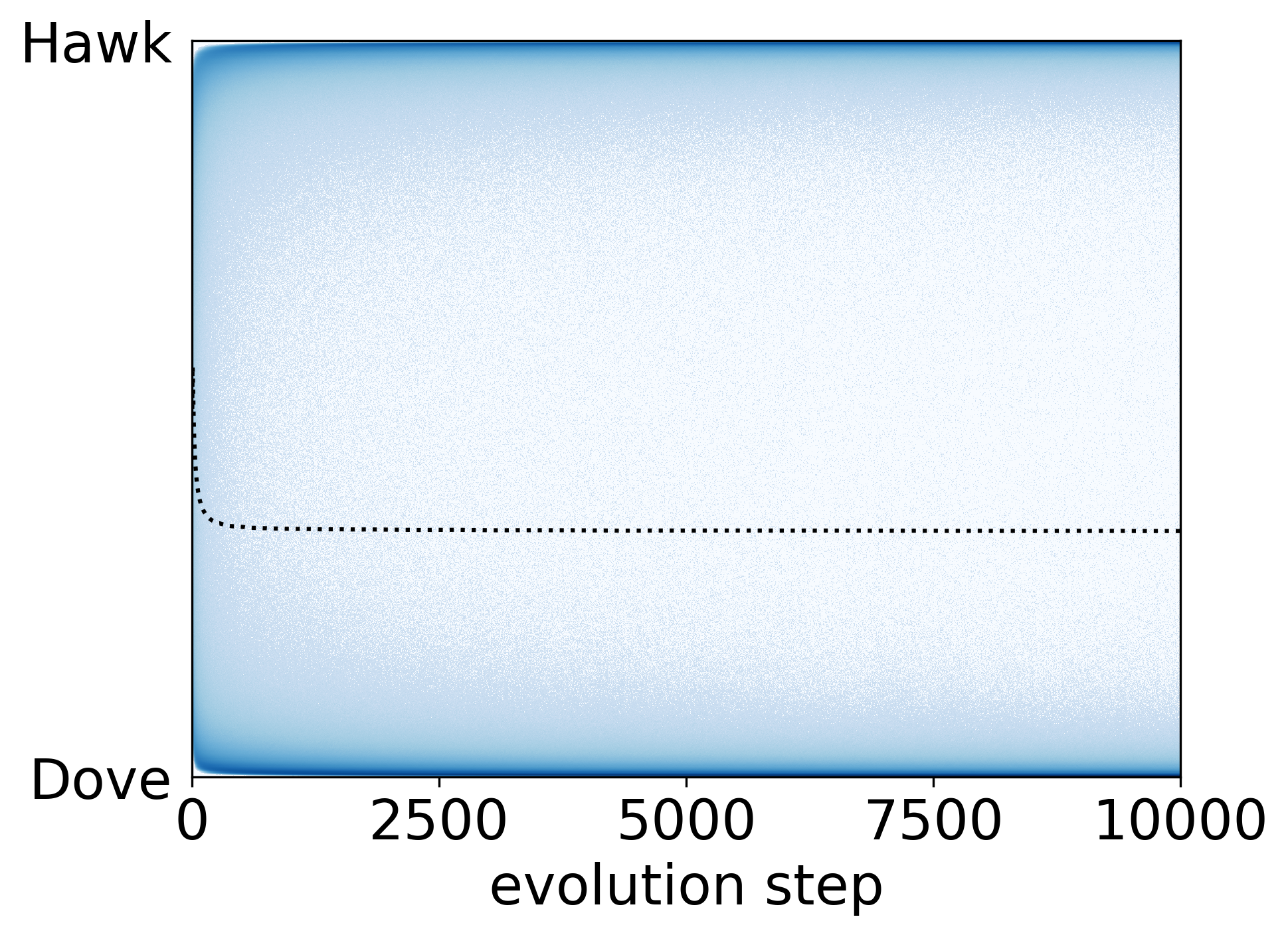}
        \caption{LOLA late evolution}
        \label{fig:hd_evo_lola2}
    \end{subfigure}
    
    \caption{Late evolution in Hawk-Dove ($f=-2$)}
    \label{fig:hd_evo}
\end{figure}

\section{Rock-Paper-Scissors}\label{ap:rps}

\begin{figure}[h]
    \centering
    \begin{subfigure}[b]{0.45\textwidth}
        \centering
        \includegraphics[width=\textwidth, trim=100 50 60 50, clip]{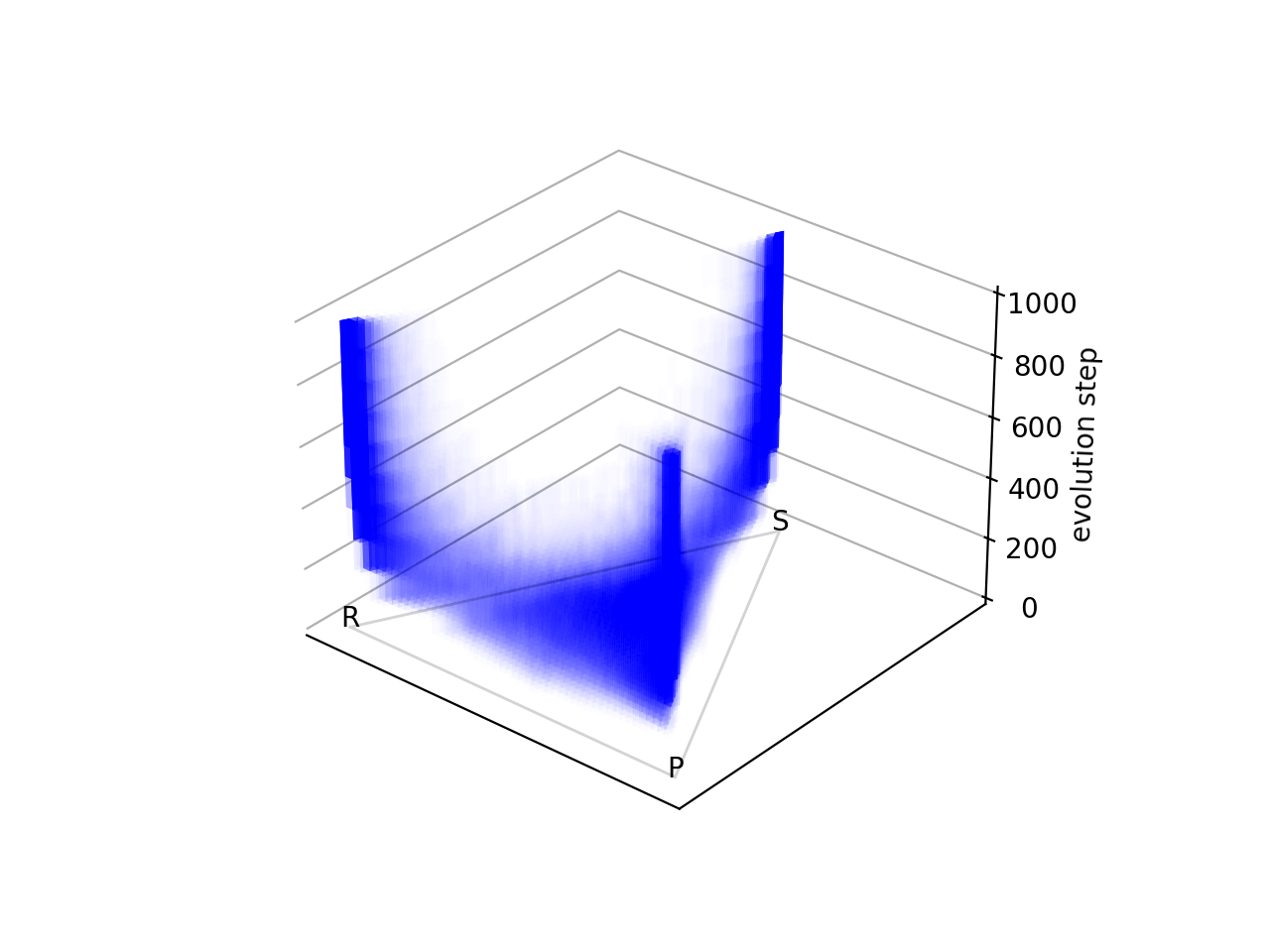}
        \caption{20\% LOLA}
        \label{fig:rps_evo_20}
    \end{subfigure}
    \hfill
    \begin{subfigure}[b]{0.45\textwidth}
        \centering
        \includegraphics[width=\textwidth, trim=100 50 60 50, clip]{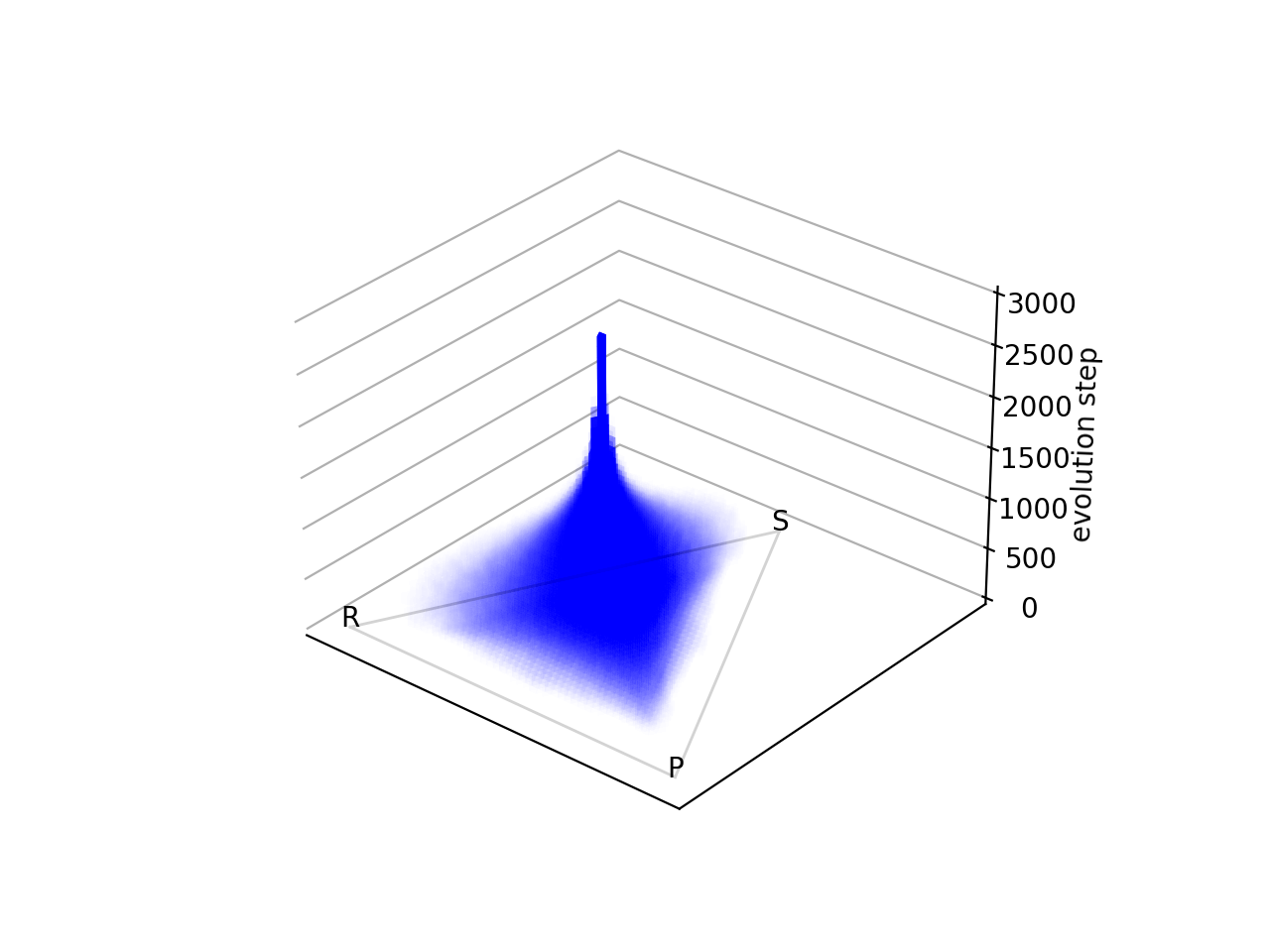}
        \caption{30\% LOLA}
        \label{fig:rps_evo_30}
    \end{subfigure}
    
    \caption{Mixed PG and LOLA in Rock-Paper-Scissors}
    \label{fig:rps_evo}
\end{figure}

\begin{figure}[h]
    \centering
    \begin{subfigure}[b]{0.32\textwidth}
        \centering
        \includegraphics[width=\textwidth, trim=0 0 0 0]{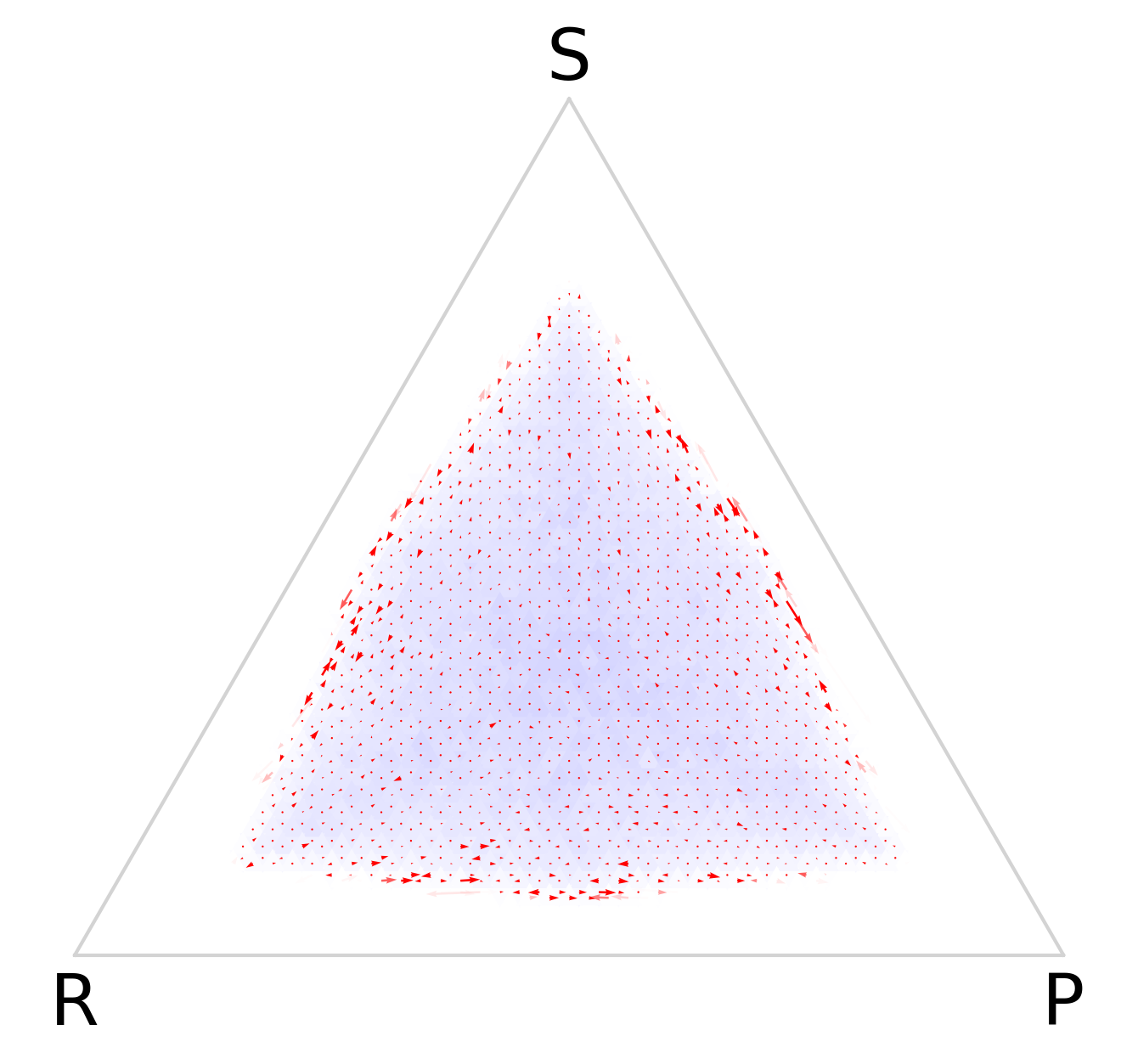}
        \caption{step 0}
    \end{subfigure}
    \hfill
    \begin{subfigure}[b]{0.32\textwidth}
        \centering
        \includegraphics[width=\textwidth, trim=0 0 0 0]{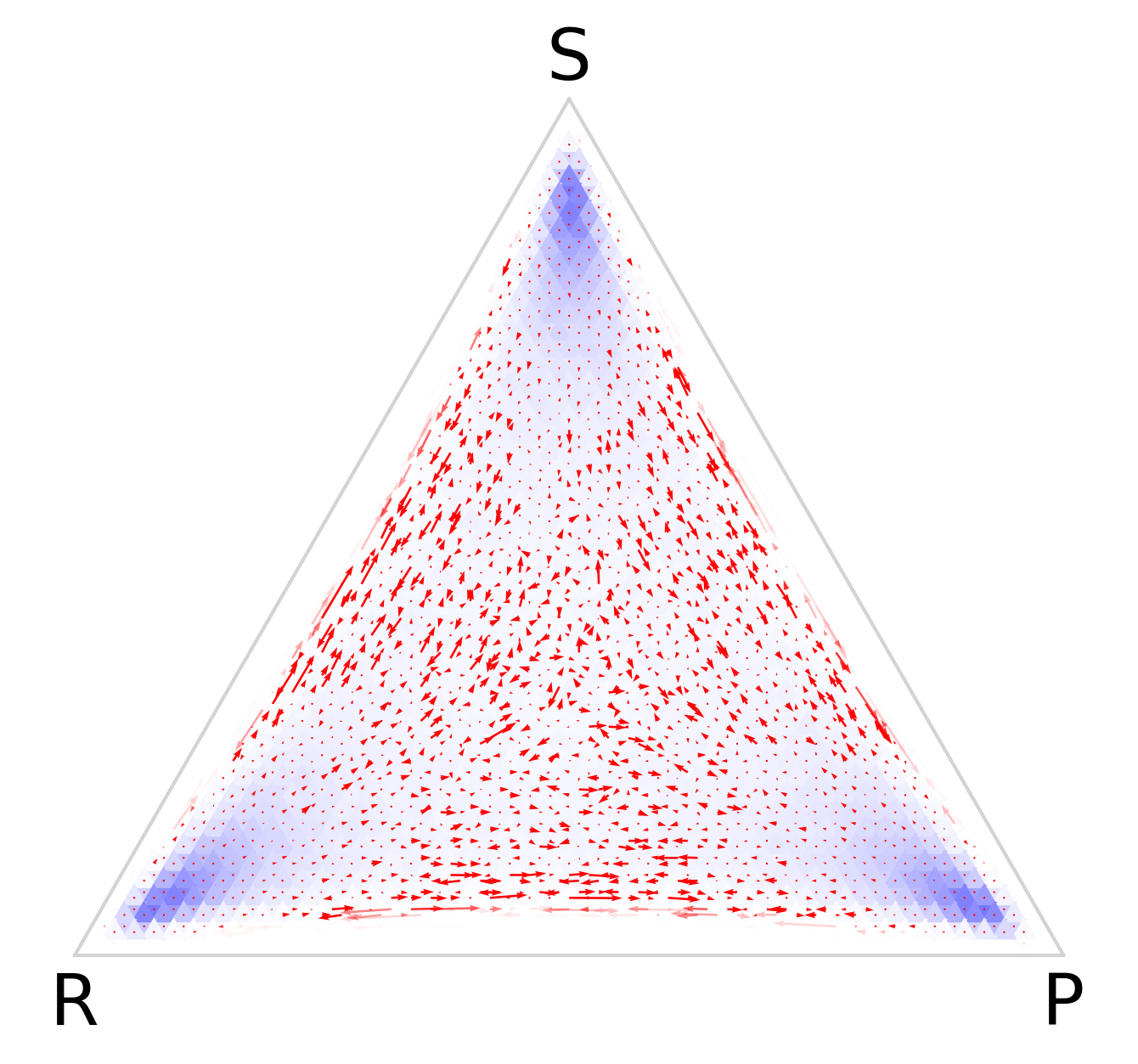}
        \caption{step 50}
    \end{subfigure}
    \hfill
    \begin{subfigure}[b]{0.32\textwidth}
        \centering
        \includegraphics[width=\textwidth, trim=0 0 0 0]{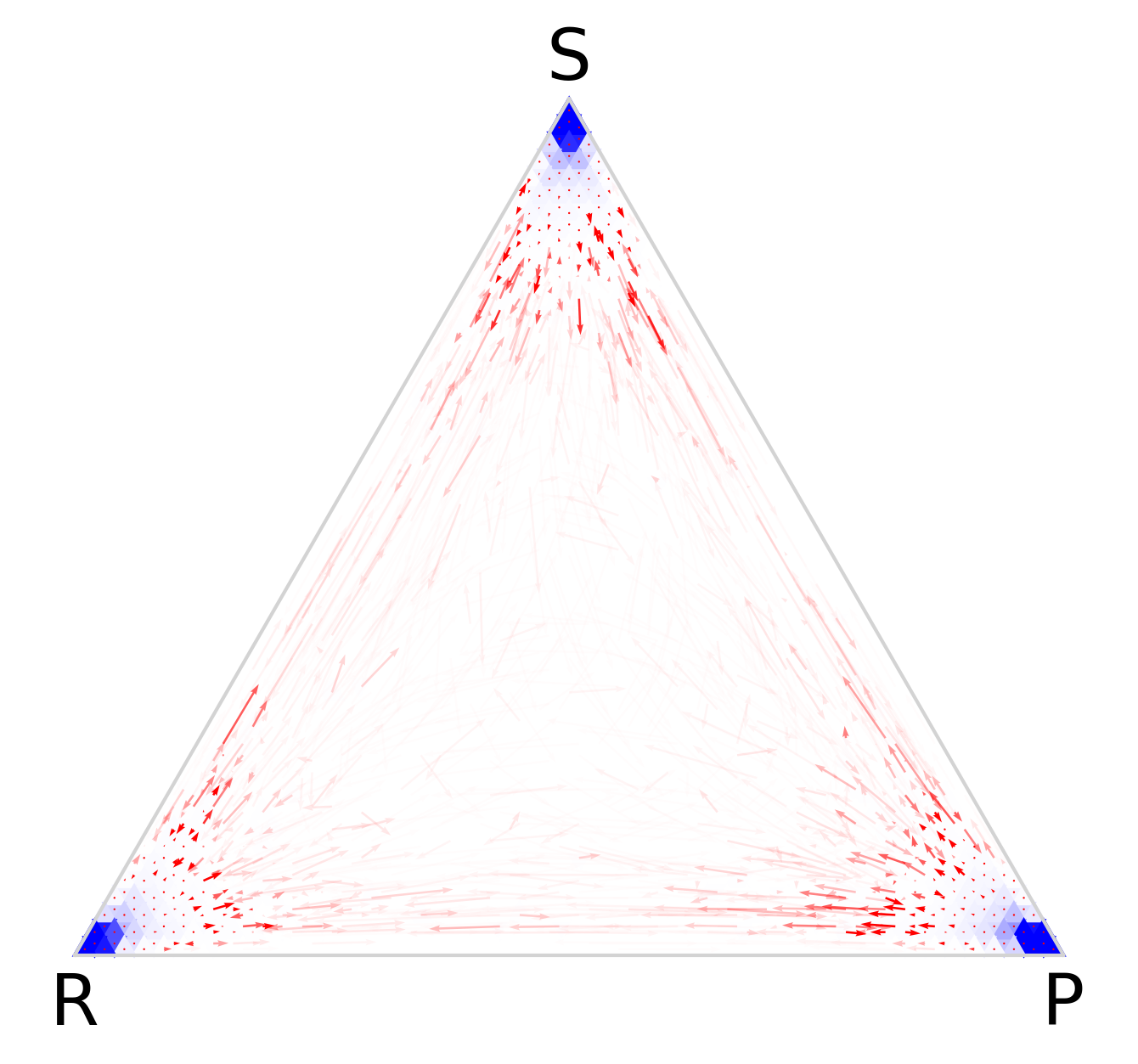}
        \caption{step 1000}
    \end{subfigure}
    \hfill
    \begin{subfigure}[b]{0.32\textwidth}
        \centering
        \includegraphics[width=\textwidth, trim=0 0 0 0]{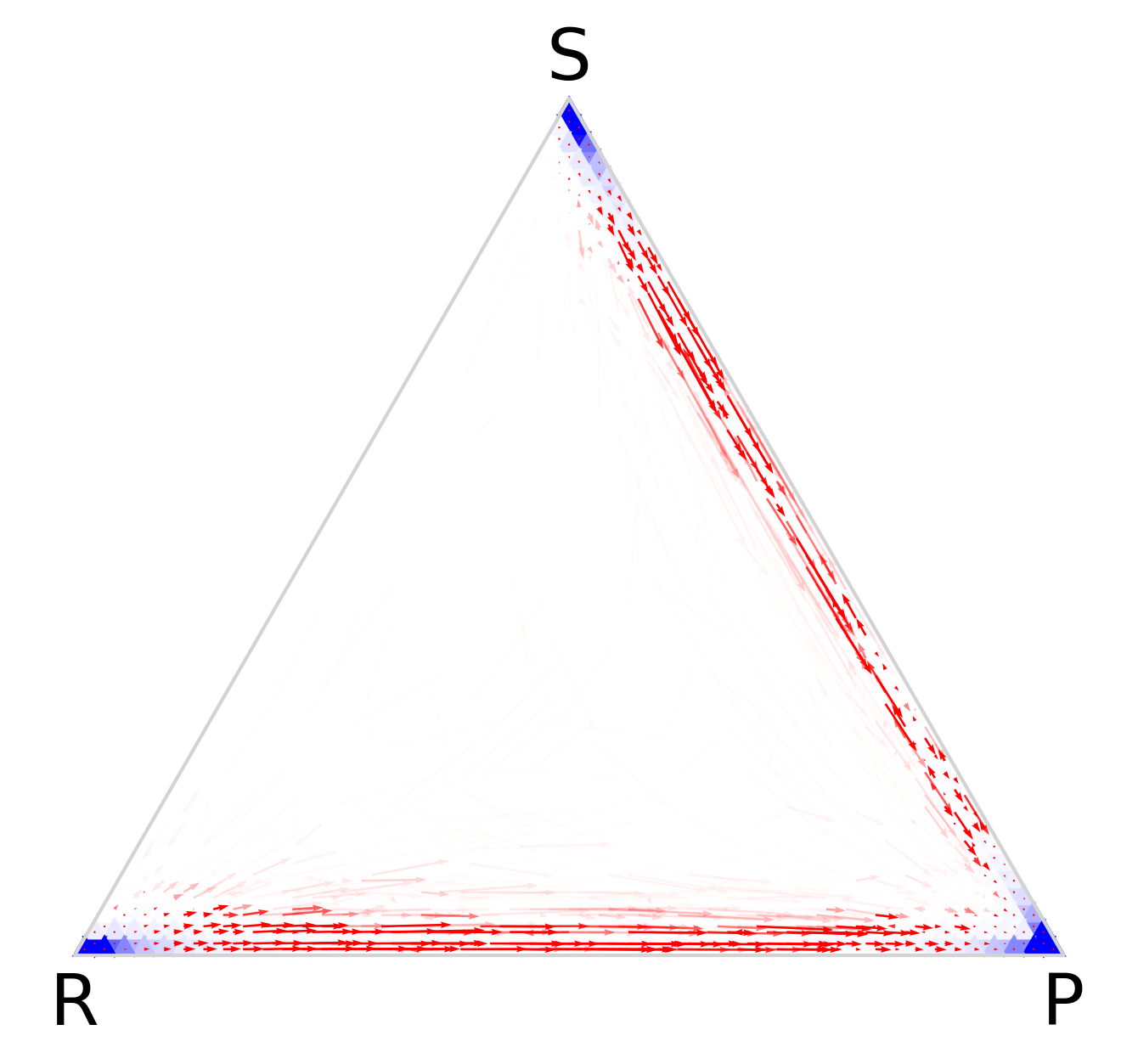}
        \caption{cycle, P attractor}
    \end{subfigure}
    \hfill
    \begin{subfigure}[b]{0.32\textwidth}
        \centering
        \includegraphics[width=\textwidth, trim=0 0 0 0]{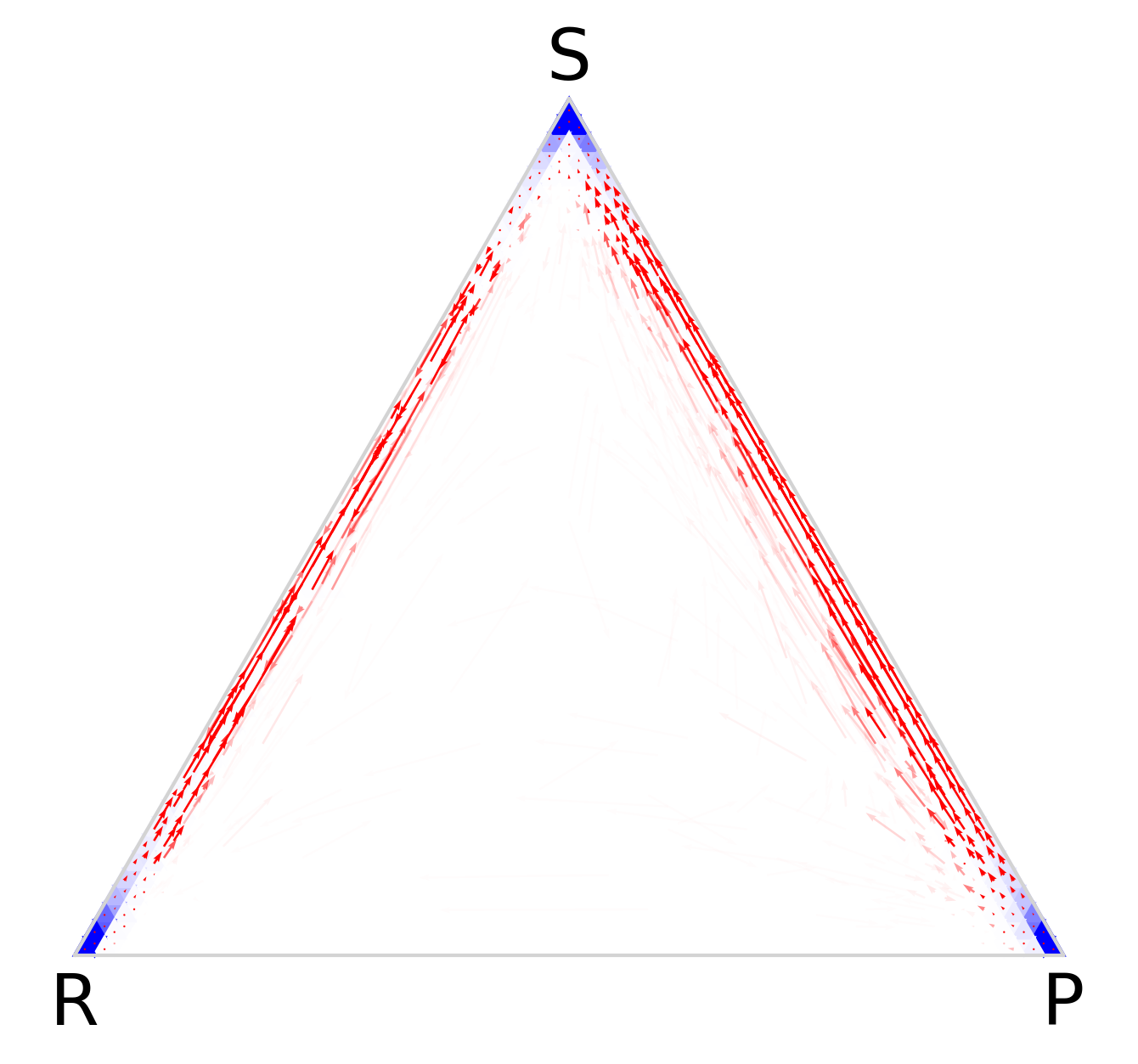}
        \caption{cycle, S attractor}
    \end{subfigure}
    \hfill
    \begin{subfigure}[b]{0.32\textwidth}
        \centering
        \includegraphics[width=\textwidth, trim=0 0 0 0]{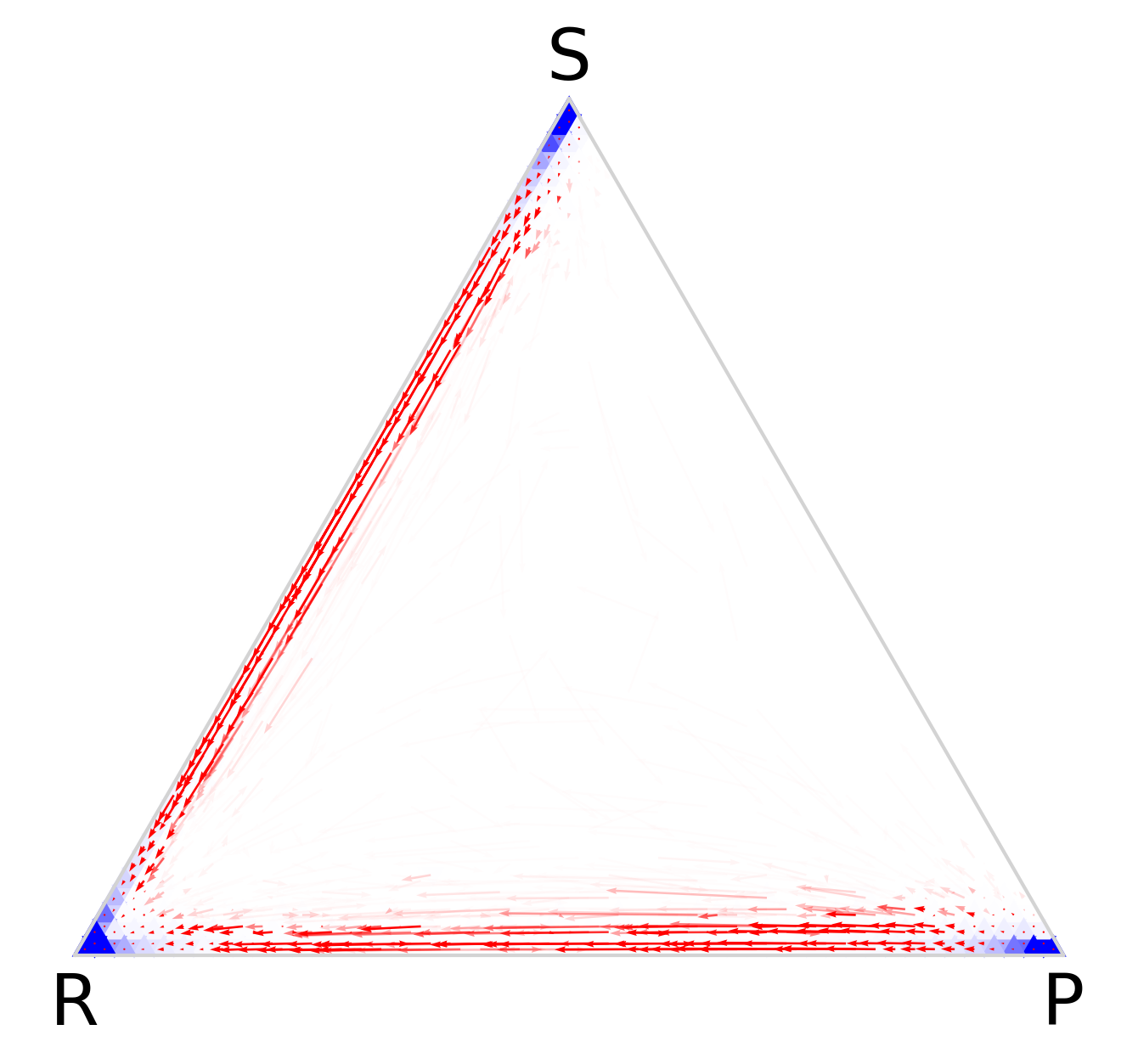}
        \caption{cycle, R attractor}
    \end{subfigure}
    
    \caption{Naive learning in RPS, late evolution. Shades of blue indicate the concentration of individuals, while red arrows indicate their average measured movement. After about 4000 evolution steps, random drift slightly unbalances the 3 groups of near-deterministic individuals, which generates a cyclic attractor pattern in the population. In the RPS model, naive learning sustains diversity.}
    \label{fig:rps_evo_pg_triangle}
\end{figure}

\begin{figure}
    \centering
    \begin{subfigure}[b]{0.32\textwidth}
        \centering
        \includegraphics[width=\textwidth, trim=0 0 0 0]{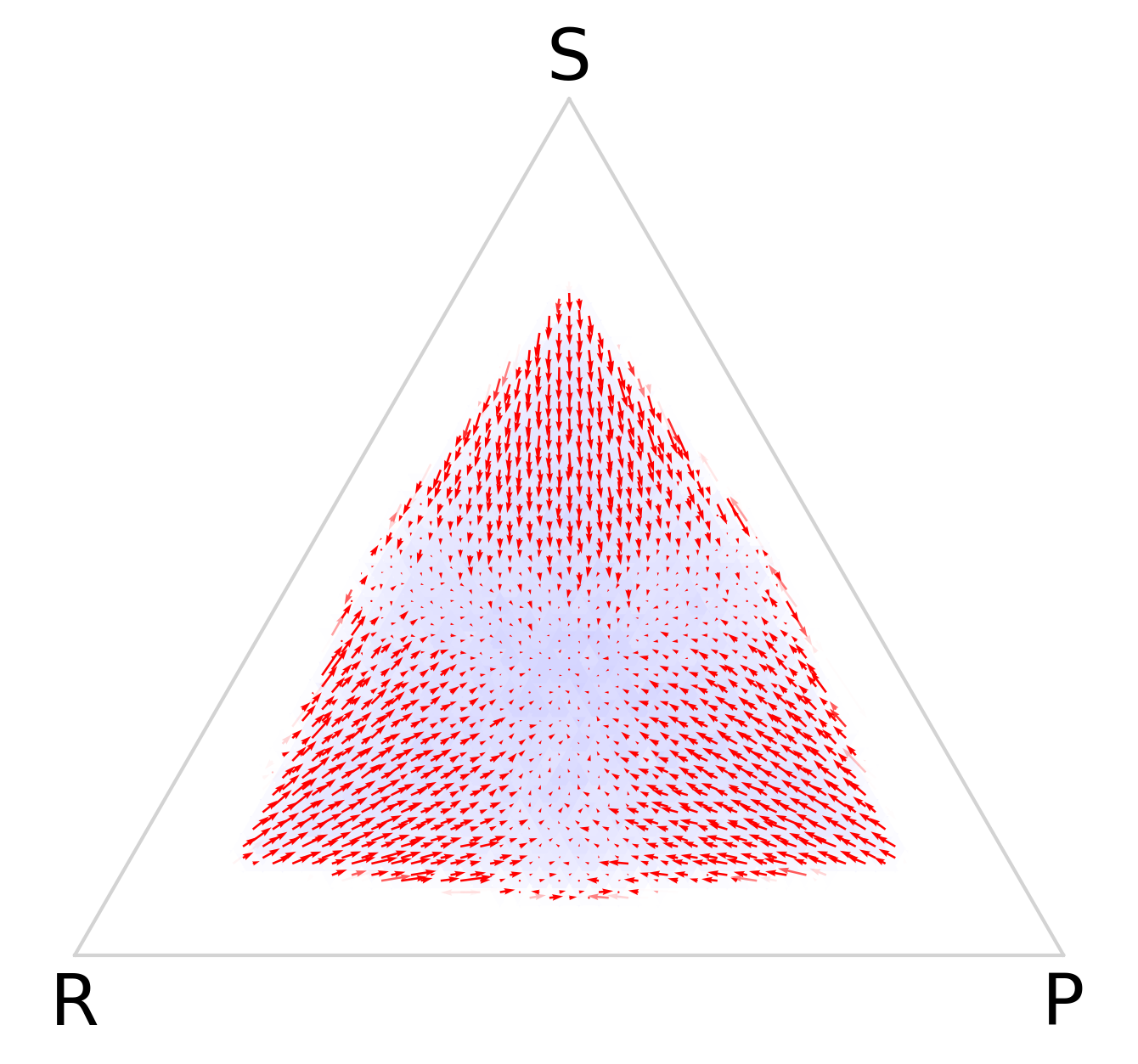}
        \caption{step 0}
    \end{subfigure}
    \hfill
    \begin{subfigure}[b]{0.32\textwidth}
        \centering
        \includegraphics[width=\textwidth, trim=0 0 0 0]{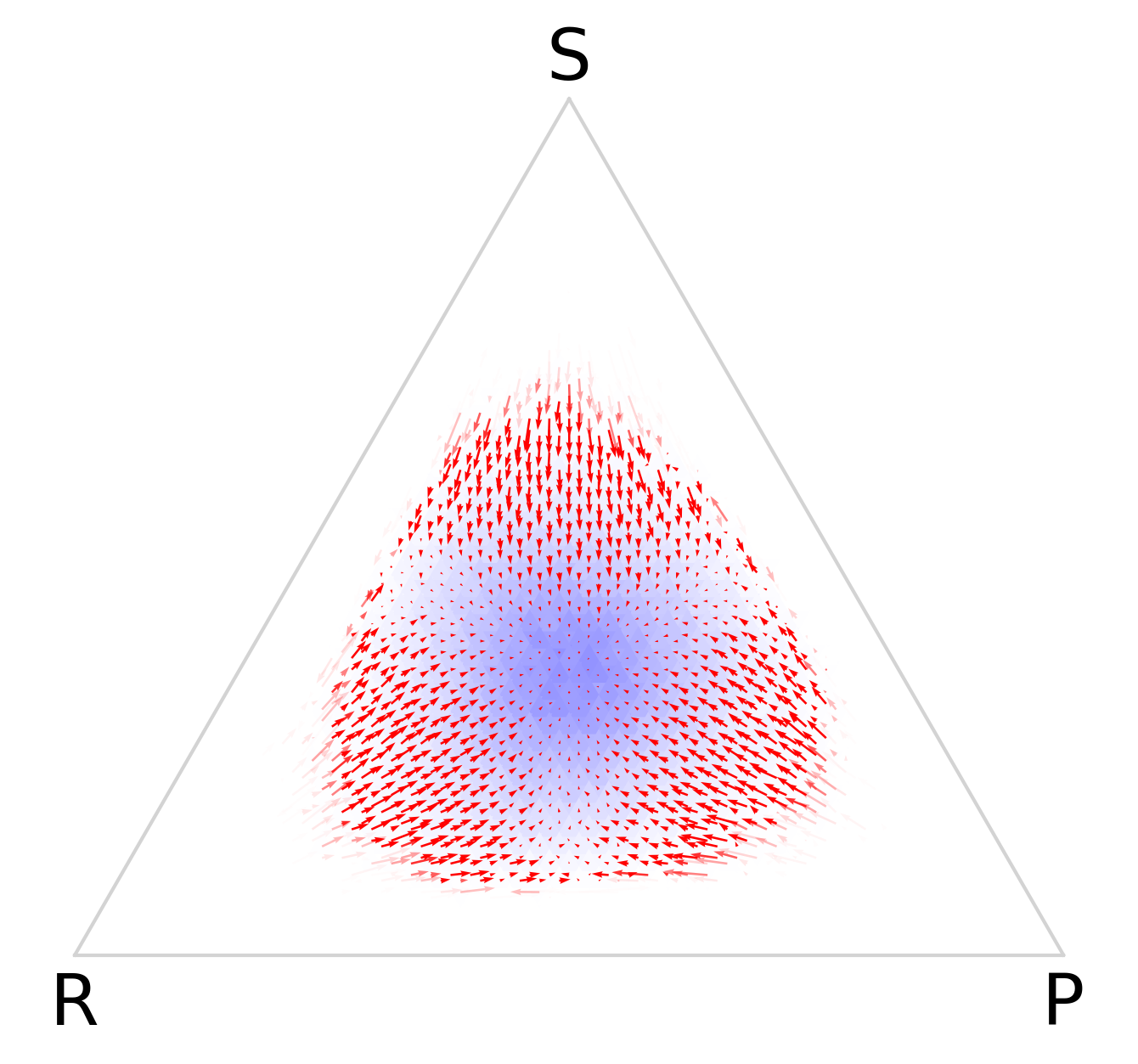}
        \caption{step 10}
    \end{subfigure}
    \hfill
    \begin{subfigure}[b]{0.32\textwidth}
        \centering
        \includegraphics[width=\textwidth, trim=0 0 0 0]{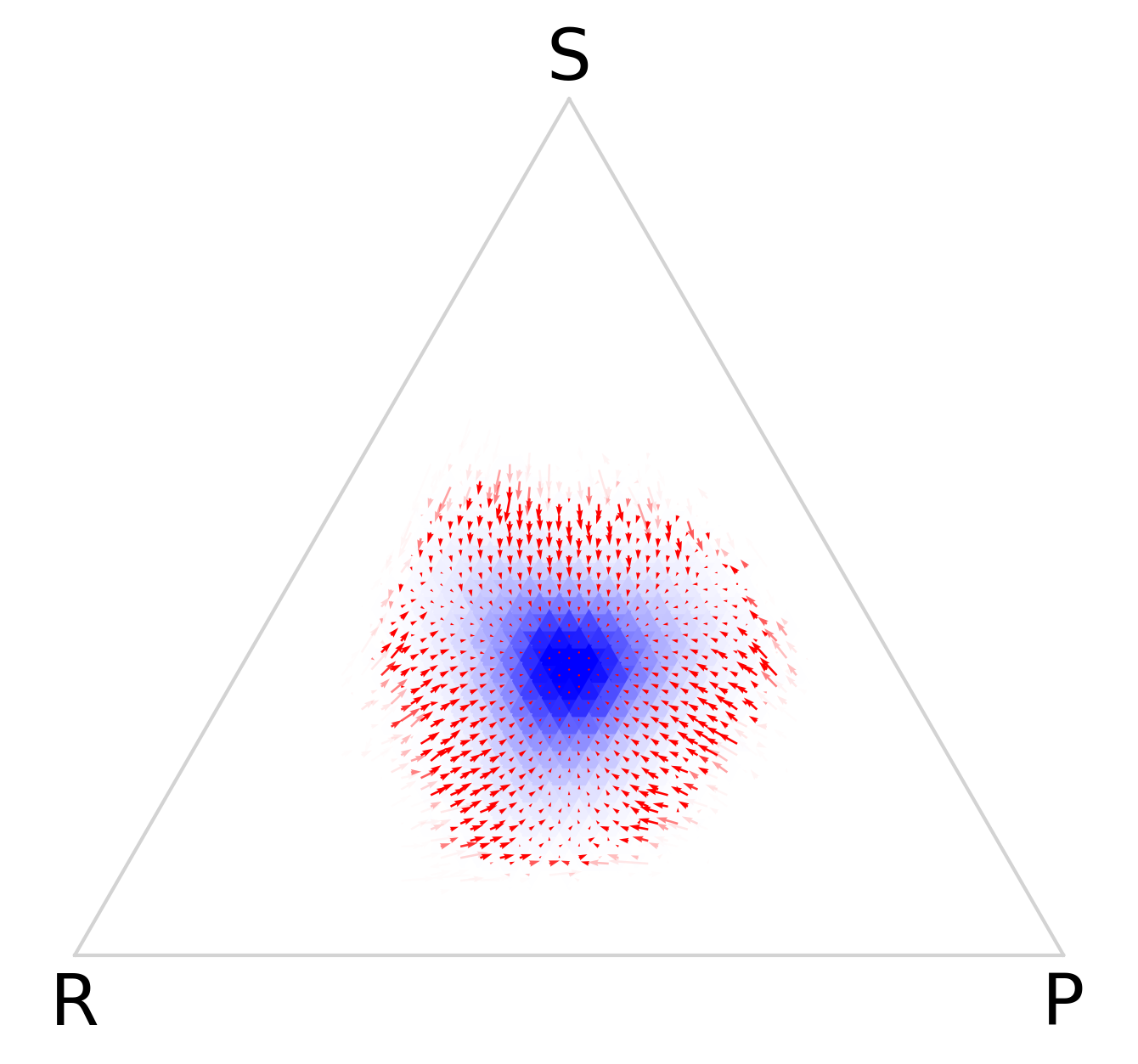}
        \caption{step 20}
    \end{subfigure}
    \hfill
    \begin{subfigure}[b]{0.32\textwidth}
        \centering
        \includegraphics[width=\textwidth, trim=0 0 0 0]{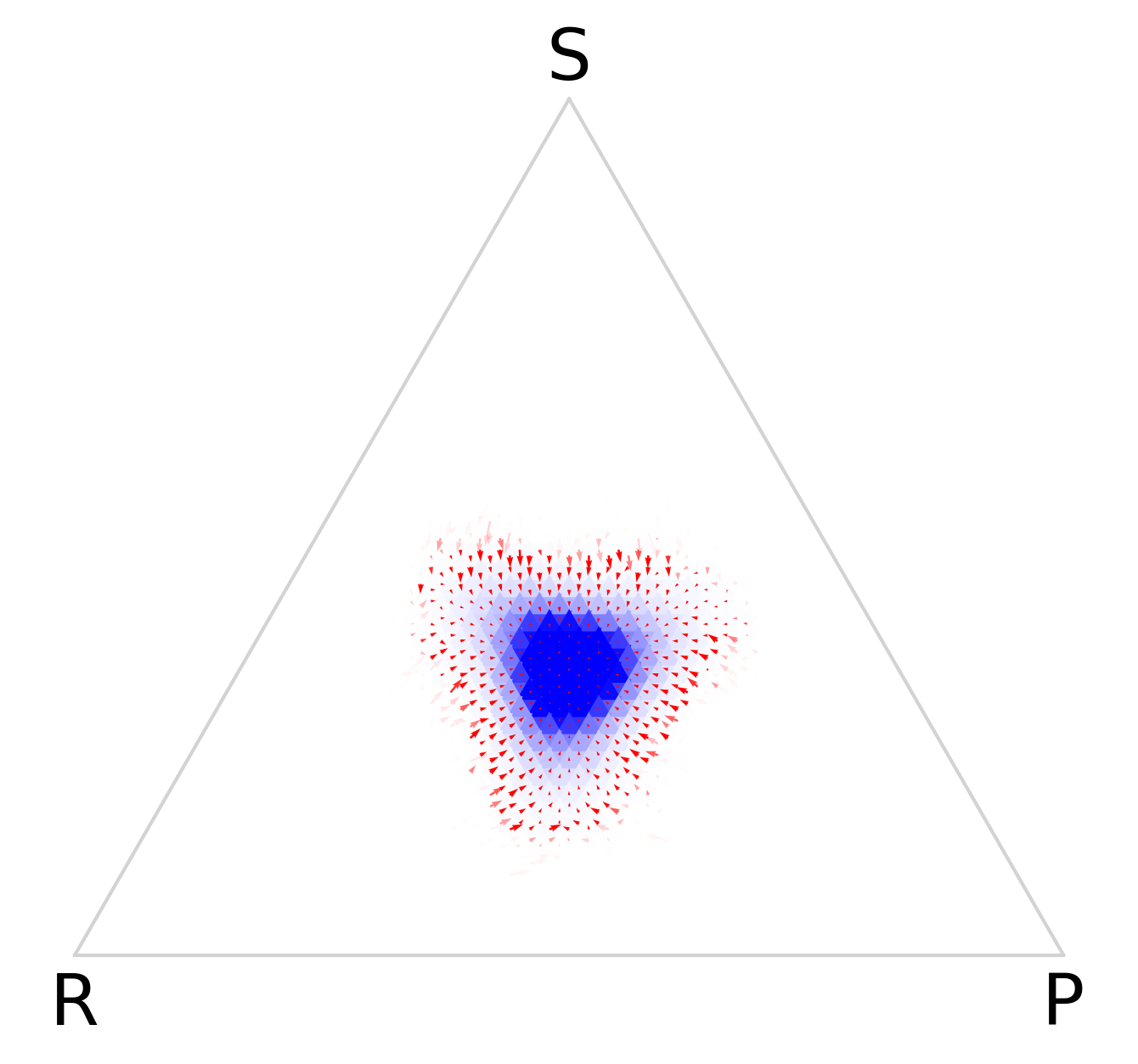}
        \caption{step 30}
    \end{subfigure}
    \hfill
    \begin{subfigure}[b]{0.32\textwidth}
        \centering
        \includegraphics[width=\textwidth, trim=0 0 0 0]{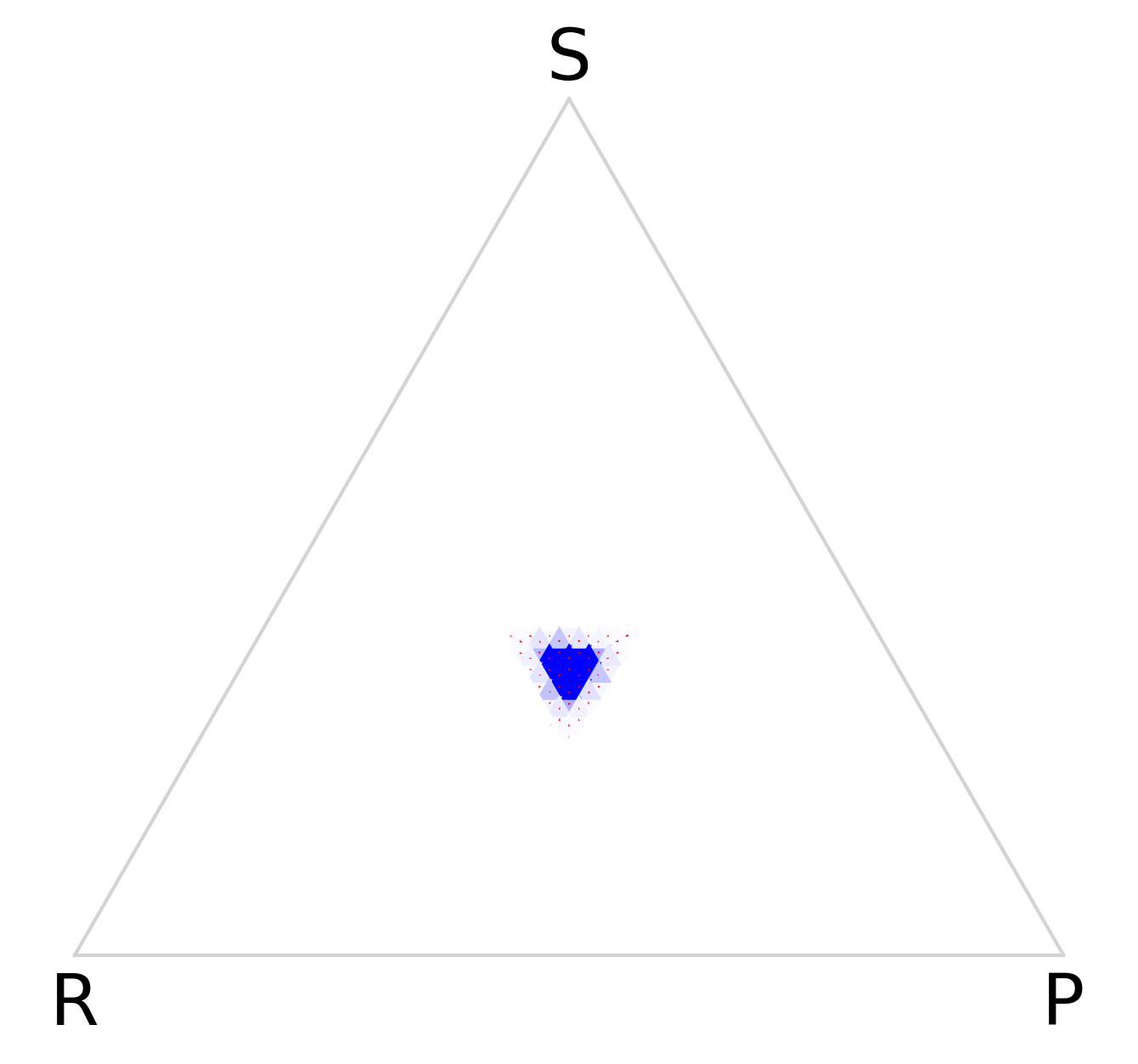}
        \caption{step 60}
    \end{subfigure}
    \hfill
    \begin{subfigure}[b]{0.32\textwidth}
        \centering
        \includegraphics[width=\textwidth, trim=0 0 0 0]{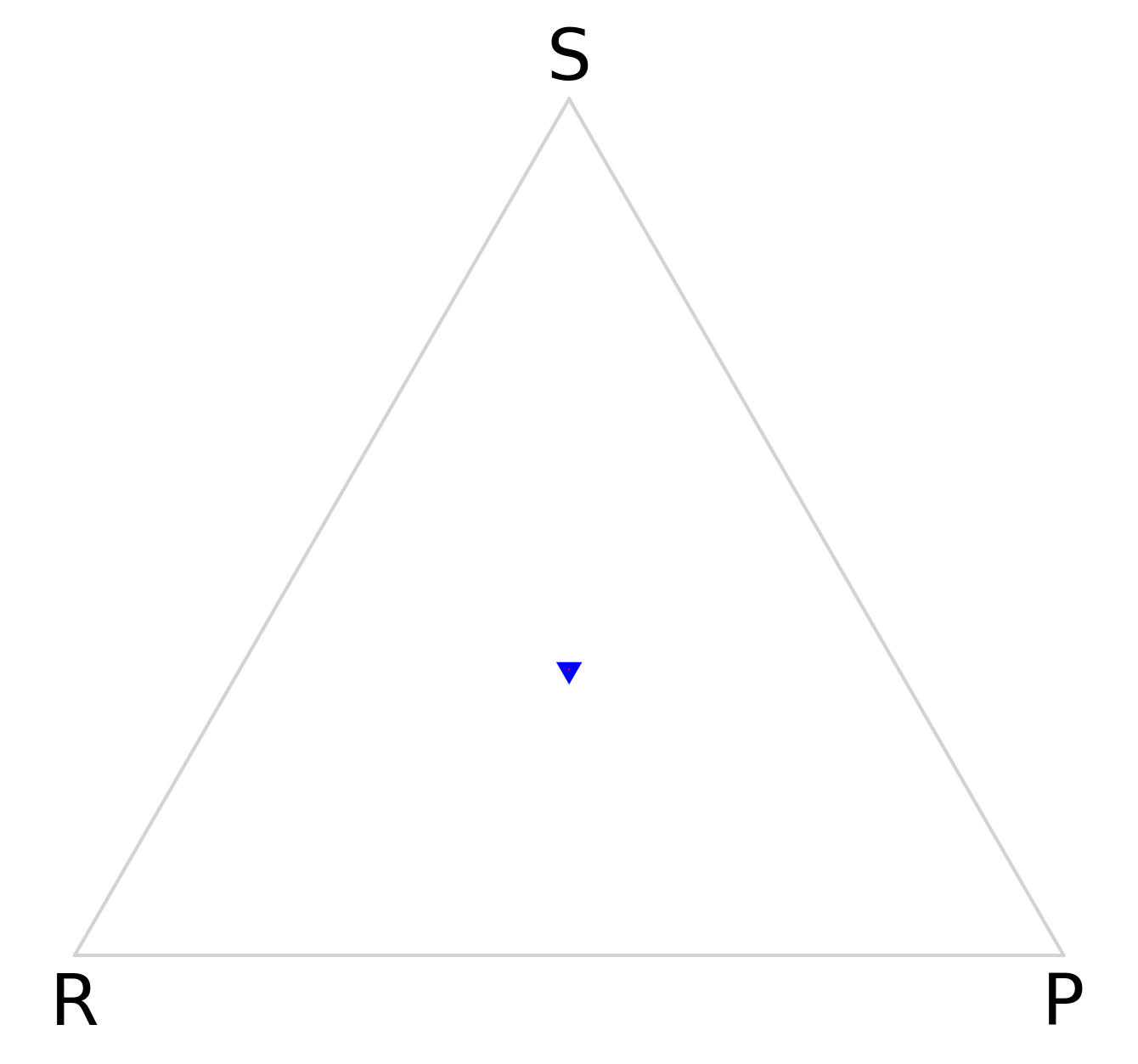}
        \caption{step 100}
    \end{subfigure}
    
    \caption{LOLA in RPS. Opponent-learning-awareness quickly brings the entire population to unanimously play the Nash equilibrium of this game (even when 70\% of the population is naive, as shown in Figure~\ref{fig:rps_evo_30}). In the RPS model, LOLA hinders diversity.}
    \label{fig:rps_evo_lola_triangle}
\end{figure}
\end{document}